\documentclass[10pt,twocolumn,letterpaper]{article}

\usepackage[pagenumbers]{cvpr} %

\usepackage[accsupp]{axessibility}  %

\usepackage{graphicx}
\usepackage{amsmath}
\usepackage{amssymb}
\usepackage{booktabs}
\usepackage{url}
\usepackage{algorithm}
\usepackage{algpseudocode}
\usepackage{longtable}

\usepackage[pagebackref,breaklinks,colorlinks]{hyperref}

\usepackage[capitalize]{cleveref}
\crefname{section}{sec.}{secs.}
\Crefname{section}{section}{sections}
\Crefname{table}{table}{tables}
\crefname{table}{tab.}{tabs.}
\Crefname{figure}{figure}{figures}
\crefname{figure}{fig.}{figs.}

\def\cvprPaperID{9021} %
\def\confName{CVPR}
\def\confYear{2023}

\newcommand{\V}[1]{\MakeLowercase{{\boldsymbol{\mathbf{#1}}}}}

\newcommand{\M}[1]{\MakeUppercase{{\boldsymbol{\mathbf{#1}}}}}

\newcommand{\T}{\top}

\newcommand{\dd}{\mathbf{d}}

\newcommand{\xx}{\mathbf{x}}

\newcommand{\bs}{\mathbf{s}}
\newcommand{\uu}{\mathbf{u}}

\newcommand{\zz}{\mathbf{z}}

\newcommand{\balpha}{\boldsymbol{\alpha}}

\newcommand{\myparagraph}[1]{\vspace{0.1em}\noindent\textbf{#1}}

\newcommand{\Optimus}{Optimus\xspace}
\newcommand{\AdafactorMLP}{Adafactor MLP\xspace}
\newcommand{\DiffPhy}{DiffPhy\xspace}

\newcommand{\myapprox}{\mathord{\sim}}

\begin{document}

\title{Transformer-Based Learned Optimization}

\author{
Erik Gärtner$^{1,2}$\thanks{Work done during an internship at Google.}
\qquad Luke Metz$^{1}$
\qquad Mykhaylo Andriluka$^{1}$
\\
\qquad C. Daniel Freeman$^{1}$ 
\qquad Cristian Sminchisescu$^{1}$
\\ \\
$^{1}$\bf{Google Research}\quad
$^{2}$\bf{Lund University}\quad \\
{\tt\small erik.gartner@math.lth.se} \\ \vspace{-2mm}
{\tt\small \{lmetz,mykhayloa,cdfreeman,sminchisescu\}@google.com}
}
\maketitle

\begin{abstract}
We propose a new approach to learned optimization that 
represents the computation of an optimizer's update step using a neural network. The
parameters of the optimizer are  learned on a set of optimization tasks with the objective to perform optimization efficiently.
Our innovation is a new neural network architecture for the learned optimizer inspired by the classic BFGS algorithm, which we call {\bf \Optimus{}}.
As in BFGS, we estimate a preconditioning matrix as a sum of rank-one updates but use a
Transformer-based neural network to predict these updates jointly with the step length and direction. 
In contrast to several recent learned optimization-based approaches \cite{metz2019understanding, metz2022practical}, our formulation allows for conditioning across the dimensions of the parameter space of the target problem while remaining applicable to optimization tasks of variable dimensionality without retraining. 
We demonstrate the advantages of our approach on a benchmark composed of objective functions
traditionally used for the evaluation of optimization algorithms, as well as on the real-world task
of physics-based visual reconstruction of articulated 3d human motion.

\end{abstract}

\begin{figure*}[tb]
  \centering
  \includegraphics[width=1.0\linewidth]{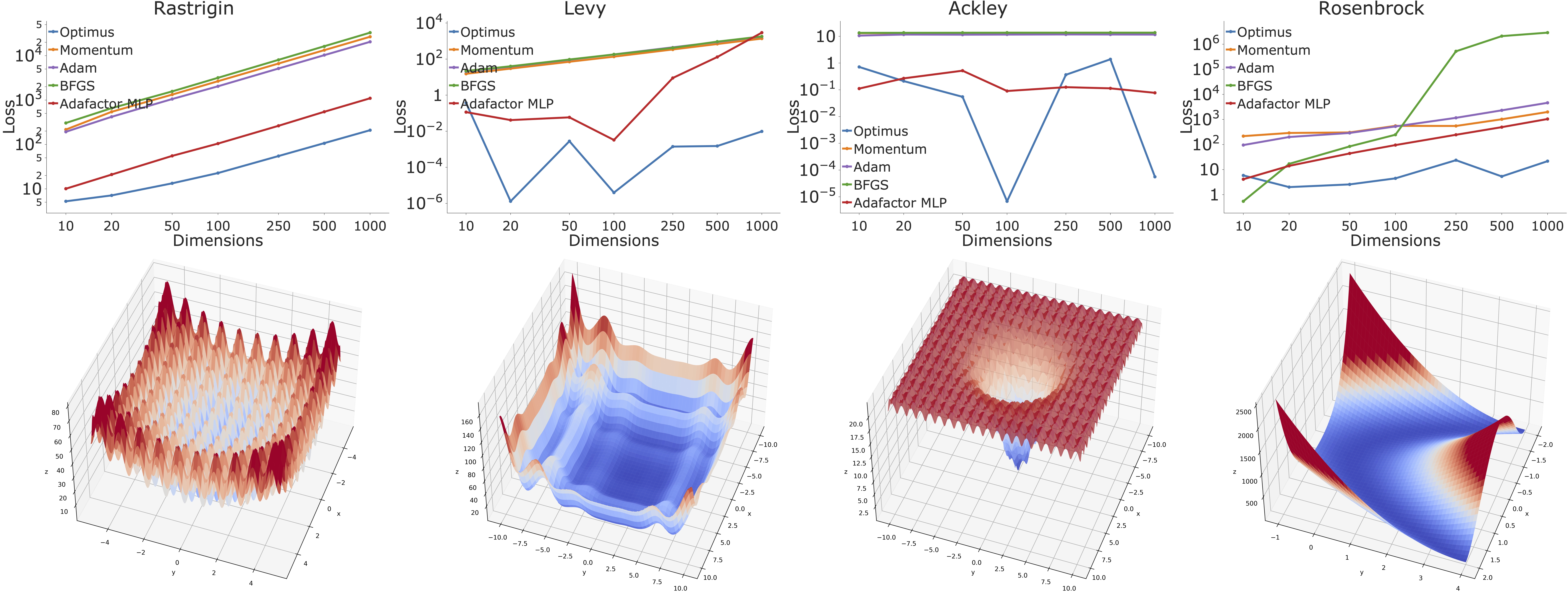}
  \caption[Examples of objective functions.]{
    Top row: Evaluation results showing average objective value reached by the optimizer for the
    corresponding objective function in the top row (y-axis) vs. dimensionality of
    the objective function (x-axis).
    Bottom row: examples of objective functions used for
    evaluation of our approach. From left to right: Rastrigin~\cite{Muhlenbein:1991:PGA},
    Levy~\cite{Laguna2005ExperimentalTO}, Ackley~\cite{ackley1987a-connectionist} and
    Rosenbrock~\cite{rosenbrock} functions. For each function, we visualize the surface of the 2d
    version.}
  \label{fig:opt_results_examples}
  \vspace{-4mm}
\end{figure*}

\section{Introduction}\label{sec:introduction}

This work focuses on a new learning-based optimization methodology.
Our approach belongs to the category of learned optimization methods, which represent the update step of an optimizer by means of an expressive function such as a multi-layer perceptron and then then estimate the parameters of this function on a set of training optimization tasks. 
Since the update function of the learned optimizers is estimated from data, it can in principle learn
various desirable behaviors such as learning-rate schedules \cite{niru2021reverse} or strategies for the exploration of
multiple local minima \cite{merchant2021icml}. 
This is in contrast to traditional optimizers such as Adam \cite{kingma2014adam}, or BFGS \cite{Flet87} in which updates are derived in terms of first-principles. However, as these are general and hard-coded, they may not be able to take advantage of the regularities in the loss functions for specific classes of problems.

Learned optimizers are particularly appealing for applications that require repeatedly solving related optimization tasks. For example, 3d human pose estimation is often formulated as a minimization of a particular loss function\cite{gartner2022diffphy, li2019cvpr,RempeContactDynamics2020,Xie_2021_ICCV}. Such approaches estimate the 3d state (\eg pose and shape) given image observations by repeatedly optimizing the same objective function for many closely related problems, including losses and state contexts. 
Traditional optimization treats each problem as independent, which is potentially
suboptimal as it does not aggregate experience across multiple related optimization runs.

The main contribution of this paper is a novel neural network architecture for learned
optimization. 
Our architecture is inspired by classical BFGS approaches that iteratively estimate the Hessian matrix to precondition the gradient.
Similarly to BFGS, our approach iteratively updates the preconditioner using rank-one updates. In contrast to BFGS, we use a transformer-based~\cite{vaswani2017attention} neural network to
generate such updates from features encoding an optimization trajectory.
We train the architecture using Persistent Evolution Strategies (PES) introduced in \cite{pmlr-v139-vicol21a}.
In contrast to prior work~\cite{metz2019understanding,metz2022practical,andrychowicz2016learning}, which update each target parameter independently (or coupled only via normalization), our approach allows for more complex inter-dimensional dependencies via self-attention and shows good generalization to different target problem sizes than those used in training.
We refer to our learned optimization approach as {\bf \Optimus{}} in the sequel.

We evaluate \Optimus{} on classical optimization objectives used to benchmark optimization
methods in the literature \cite{Laguna2005ExperimentalTO,simulationlib,rosenbrock} (cf.~fig.~\ref{fig:opt_results_examples})
as well as
on a real-world task
of physics-based human pose
reconstruction. In our experiments, we typically observe that \Optimus{} is able to
reach a lower objective value compared to popular off-the-shelf optimizers while taking fewer
iterations to converge. For example, we observe at least a 10x reduction in the number of update steps
for half of the classical optimization problems (see \cref{fig:opt_results_classical_rel_iters}). 
To evaluate \Optimus{} in the context of physics-based human motion reconstruction, we apply it in conjunction with \DiffPhy{}, which is a differentiable physics-based human model introduced in \cite{gartner2022diffphy}. 
We experimentally demonstrate that \Optimus{} generalizes well across diverse human motions (\eg from training on walking to testing on dancing), is notably (5x) faster to meta-train compared to prior work \cite{metz2022practical}, leads to reconstructions of better quality compared to BFGS, and is faster in minimizing the loss.

\section{Related Work}\label{sec:discussion}
Learned optimization is an active area of research, and we refer the reader to an excellent tutorial~\cite{amos2022tutorial} and survey~\cite{chen2021learning} for a comprehensive review of the literature. Our approach is generally inspired by \cite{andrychowicz2016learning, wichrowska2017learned, ke17iclr, metz2019understanding} and is most closely related to \AdafactorMLP{}
\cite{metz2022practical}.
One of the distinguishing properties of \Optimus{}
compared to \AdafactorMLP{} is the ability to couple optimization updates along different dimensions. Arguably coupling of dimensions can be added to \AdafactorMLP{} through additional features such as radial
features from~\cite{merchant2021icml} that capture pairwise interactions between dimensions. In contrast, the advantage
of \Optimus{} over such extensions is that dimension coupling is learned from data and is not
limited to be pairwise. Other work incorporates conditioning in some aggregated space. For example, both \cite{wichrowska2017learned} and \cite{metz2020tasks} introduce hierarchical conditioning mechanisms that operate on individual layers and in a global setting.
The \Optimus{} approach can be seen as constructing a learned preconditioner to account for the underlying cost function curvature.
The use of meta-learned curvature has been explored in \cite{park2019meta}, though only in the context of few-shot learning strategies.

There exist prior work on using learned optimization in the human pose estimation literature
\cite{song2020human} as well as approaches that iteratively refine the solution in an optimization-like
fashion based on recurrent neural networks \eg \cite{zanfir2020neural,carreira16cvpr}.
The work of \cite{song2020human} is perhaps the most similar to ours in that it genuinely employs a learned optimization in the framing described in \cref{sec:overview}.
However, \cite{song2020human} applies it to the simpler problem of monocular 3d pose estimation. On that task 
their method converges in as few as four iterations, thus making meta-training based on stochastic gradient descent feasible.
By design, the approach of \cite{song2020human} does not generalize to optimization instances with variable (thus different) dimensionality with respect to training, as they employ a single MLP that predicts the entire
update vector. Moreover, their work has yet to be evaluated for more complex tasks such as those considered in this paper.

\section{Overview and Background}\label{sec:overview}

\begin{figure*}[tb]
  \centering
   \includegraphics[width=1\linewidth]{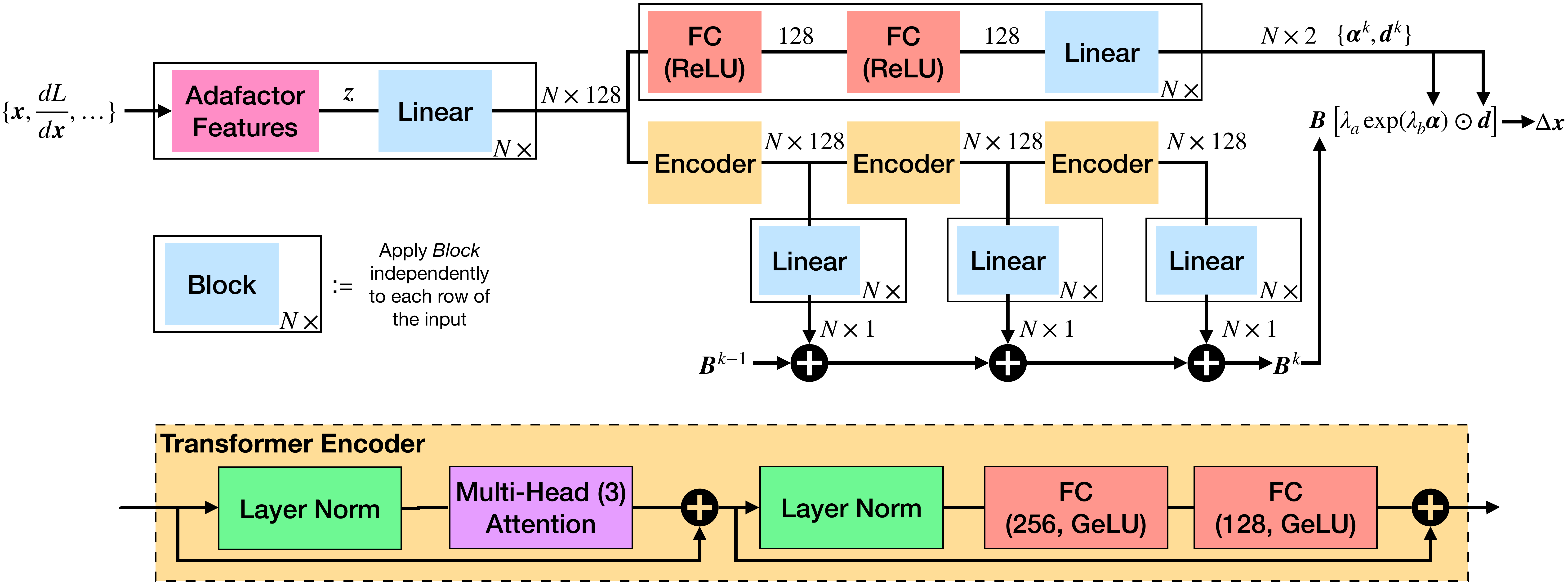}
   \vspace{-4mm}
  \caption{Schematic overview of applying our Transformer-based learned optimizer, \emph{Optimus}. The architecture of Optimus consisting of $L = 3$ stacked Transformer encoders that predict rank one updates to a learned pre-conditioning matrix $\M B$ and a separate branch that predicts step size $\V a$ and step direction $\V d$. The network uses the \emph{Adafactor MLP} features introduced in \cite{metz2022practical} as input. Note: to allow our architecture to generalize to function of different dimensions all \emph{linear} and \emph{FC} layers are applied per-parameter.}
   \vspace{-2mm}
  \label{fig:optimus_network}
\end{figure*}

In this paper, we leverage a general approach to learned
optimization as introduced in \cite{bengio1990learning, bengio1992optimization,andrychowicz2016learning,metz2019understanding}, which we review in \cref{sec:learned_optimization}.
Equipped
with this background, we then introduce the details of our new \Optimus{} architecture for learned
optimization in \cref{sec:methodology}. Then in \cref{sec:experiments} we present experimental
results comparing \Optimus{} to prior work in learned optimization~\cite{metz2019understanding} and
against standard off-the-shelf optimizers.

\subsection{Learning an Optimizer}\label{sec:learned_optimization}

Learned optimizers are a particular type of meta-learned system which commonly uses a neural network
to parameterize a gradient-based step calculation that can then be used to optimize some objective function~\cite{andrychowicz2016learning}. 
To demonstrate this class of models, let us consider an optimization problem
$\text{argmin}_{\xx} L(\xx)$. 
Gradient-based optimization algorithms such as gradient descent (GD) aim to solve the problem by iteratively modifying
the parameters $\xx$ using an update function $U$ which takes gradients of $L$ along the optimization trajectory as input
$\xx_{k+1} = \xx_{k} - U(\nabla_{\leq k} L(\xx_{1:k}))$, where $\xx_k$ are the parameters 
at  step $k$
and $\nabla_{\leq t}L(\xx_{1:k}) = \{\nabla L(\xx_1), \ldots, \nabla L(\xx_k)\}$.
For example, in the case of GD, the update function is simply 
$U_{\text{gd}}(\nabla_{\leq k}L(\xx_{1:k})) = -\alpha\nabla L(\xx_k)$,
where $\alpha$ is a learning rate hyperparameter.
A learned optimizer is a particular type of update function, which itself is parameterized by a set of meta-parameters $\V \theta$, and with
possibly more features (\eg, the current parameter values).
Then, as with GD, it can be iteratively applied to improve the loss.

In this paper, we build on learned optimization as proposed in
\cite{metz2019understanding, metz2022practical, pmlr-v139-vicol21a}.
In that approach, the update function
$U=U(\zz|\V \theta)$ is parameterized based on a small multilayer perceptron (MLP) with weights $\V \theta$, which is applied
independently to each dimension of a feature vector $\V z$. For each parameter, the update function takes a vector of
features $\zz$ as input, including gradient information, as well as additional features such as
exponential averages of squared past gradients as done in Adam~\cite{kingma2014adam} or RMSProp~\cite{tieleman2012lecture}, momentum at multiple timescales~\cite{lucas2018aggregated}, as well as factored features inspired by Adafactor~\cite{shazeer2018adafactor}.
We refer to this approach as \AdafactorMLP{} in the sequel.
Training an \AdafactorMLP{} optimizer amounts to minimizing a meta-loss with respect to
parameters $\V \theta$ on a meta training-set of optimization problem instances. 
The meta-loss is given by $\sum_{k}L(\xx_{k})$ where the sum runs over the parameter states of the optimization trajectory.
Minimizing the meta-loss is often implemented via truncated backpropagation through unrolled
optimization trajectories~\cite{werbos1990backpropagation, maclaurin2015gradient, andrychowicz2016learning, wichrowska2017learned}.
As discussed in \cite{metz2019understanding, metz2021gradients, pmlr-v139-vicol21a} typical meta-loss surfaces are noisy
and direct gradient-based optimization is difficult due to exploding gradients. 
To address this, we minimize the smoothed version of a meta-loss as in~\cite{metz2019understanding} using Adam~\cite{kingma2014adam} and adopt Persistent Evolution Strategies (PES)~\cite{pmlr-v139-vicol21a} to compensate for bias due to truncated back-propagation~\cite{wu2018understanding}.

\section{Our Approach}\label{sec:methodology}

Our transformer-based learned optimizer, \emph{Optimus}, is inspired by the BFGS~\cite{Flet87} rank-one approximation approach to estimating the inverse Hessian, which is applied as a preconditioning matrix in order to obtain the descent direction. The parameter update is the product of a descent direction ($\V s^k$) produced by a learned optimizer that operates on each parameter independently, and a learned preconditioner ($\M B^k$) where $\M B^k$ is an $N \times N$ matrix which supports conditioning over the entire parameter space. We update $\M B^k$ with $L$ rank-one updates on each iteration. The full update is thus given by
\begin{equation}\label{eq:optimus_update}
  \Delta \V x^{k} = \M B^k \V s^k \text{,}
\end{equation}
where $\Delta \V x^k$ is the parameter update at iteration $k$. See  \cref{fig:optimus_network} for an overview. 

\myparagraph{Per-Parameter Learned Descent Direction ($\V s^k$).}
Let us denote a feature vector describing the optimization state of the $n$-th parameter at iteration $k$ as $\V z_n^k$.
As in \cite{metz2022practical} we predict per-parameter updates using a simple MLP that takes the feature vector $\V z^k_n$ as input and outputs a log learning rate $\alpha^k_n$ and update direction $d^k_n$ that are combined into a per-parameter update as
\begin{equation}\label{eq:adafactor_mlp_step}
s^k_n = \lambda_a\exp(\lambda_b\alpha^k_n)d^k_n,
\end{equation}
where $\lambda_a=0.1$, and $\lambda_b=0.1$ are hyperparameters which are constant throughout meta-training. 
Note that at that stage we \emph{independently} predict the update direction and magnitude for each dimension of the vector $\V x$. In particular, the MLP weights are shared across all the dimensions of $\V x$. We use a small 4-layer MLP with 128 units per layer at that stage and did not observe improvement when with larger models (see tab.~\ref{tab:model_ablation}). We use the same features $\V z^k_n$ as \cite{metz2022practical} and similarly to \cite{metz2022practical} normalize the features to have a second moment of $1$. We include the feature list in the supplementary material for completeness. 

\myparagraph{Learned Preconditioning ($\M B^k$).}
Next, we introduce a mechanism to couple the optimization process of each dimension of $\V x$ and enable the optimization algorithm to store information across iterations. Intuitively such coupling should lead to improved optimization trajectories by capturing curvature information, similar to how
second-order and quasi-Newton methods improve over first-order methods such as gradient descent.
We define these updates as a low-rank update followed by normalization:
\begin{equation*} \label{eq:optimus}
\tilde{\M B}^{k+1} = \M B^{k} + \sum_{l=1}^{L} \uu^{k}_l(\uu^{k}_l)^\T,  \quad \M B^{k+1} = \tilde{\M B}^{k+1} \mathbin{/} \|\tilde{\M B}^{k+1}\| \text{,}
\end{equation*}
where we initialize with $\M B^{0} = \M I^{N\times N}$.
To predict the $N$-dimensional vectors $\V u^{k}_l$ we apply a stack of $L$ Transformer encoders \cite{vaswani2017attention} to a set of per-parameter features linearly mapped to $d=128$ dimensions. 
Note that traditionally Transformer architecture has been applied to sequential data, whereas here we use it to aggregate information along the parameter dimensions. 
We visualize the architecture in~fig.~\ref{fig:optimus_network} for the case $L=3$. 
The $i$-th element of $\V u^k_{l}$ is computed by applying a layer-dependent linear mapping $\M M_l$ to the \mbox{$i$-th} row of the output $\M E^{kl}$ of the Transformer encoder at the layer $l$: $\V u^k_{li} = \M M_l(\M E^{kl}_{i:})$.

Note that our formulation of the update equations for $\M B^k$ supports several desirable properties. First, it enables coupling between updates of individual parameters through the self-attention of the encoders. Secondly, our formulation does so without making the network specific to the objective function dimensions used during training. This allows us to readily generalize to problems of different dimensionality (see \cref{sec:expclassical}). Finally, it allows the
optimizer to accumulate information across iterations, similarly to how BFGS~\cite{Flet87}
incrementally approximates the inverse Hessian matrix as optimization goes on. Effectively our methodology works by learning a \emph{preconditioning} for the first-order updates estimated in other learned optimizers, such as the \AdafactorMLP{}~\cite{metz2022practical}. While this preconditioner considerably increases the step quality, its computational cost grows quadratically in the number of parameters.

\myparagraph{Stopping Criterion.} During meta-training, we unroll the optimizer for $50$ steps, but at test time, we run \Optimus{} using a stopping criterion based on relative function value decrease, as in classical optimization. We terminate the search if $f(\V x^k) > \frac{1}{N}\sum_{i=1}^{N}\beta f(\V x^{k-i}) + \epsilon$, \ie if  the function value at step $k$ is greater than the average function value in the previous $N$ steps. We do not apply this criterion for the first $N$ steps. We set $N=5$ in our experiments.

\section{Experiments}\label{sec:experiments}
We evaluate our learned optimization methodology on two tasks. We first present results of a
benchmark composed of objective functions typically used for evaluation of optimization methods, and
then present results for articulated 3d human motion reconstruction in \cref{sec:physics_based_hpe}.

\myparagraph{Baselines.} We compare the performance of our \Optimus~optimizer to standard
optimization algorithms BFGS~\cite{Flet87}, Adam~\cite{kingma2014adam}, %
and gradient descent with momentum (GD-M).  We independently tune the learning rate of Adam and
GD-M for each optimization task given by objective function and input dimensionality using
grid-search.  To that end we test $100$ candidate learning rates between $10^{-6}$ and $1$ and choose
the learning rate that results in lowest average
objective value after running optimization for $64$ random initializations.  Finally, we also
compare our approach to the state-of-the-art learned optimizer \AdafactorMLP
\cite{metz2019understanding,metz2022practical} using the publicly available
implementation\footnote{\url{https://github.com/google/learned_optimization}}.

\subsection{Standard evaluation functions}\label{sec:expclassical}
\myparagraph{Evaluation benchmark.}  We define a benchmark composed of $15$ objective
functions frequently used for evaluation of optimization algorithms. We show a few examples of such
functions in \cref{fig:opt_results_examples} and provide a full list in the supplementary
material. To define the benchmark we use the catalog of objective functions available at
\cite{simulationlib}, focusing on the functions that can
be instantiated for any dimensionality of the input. We include both seemingly easy to optimize
functions (\eg ``Sphere'' function\footnote{$f_{\mbox{sphere}}(x) = \sum_{i=1}^dx_{i}^2$, where $x\in\mathbf{R}^d$}) as well as more challenging functions with multiple local minima (\eg Ackley function
\cite{ackley1987a-connectionist}) or difficult to find global minima located at the bottom of an
elongated valley (\eg Rosenbrock function \cite{rosenbrock}). We use versions of these functions
with input dimensionality of $2-100$ for training, and then evaluate on the dimensions $250$, $500$
and $1000$ to show that optimizer can generalize to different dimensionality of the input. This
gives us a test set of $45$ objective functions to evaluate on. To prevent the learned optimizer
from memorizing the position specific information we add random offsets to the objective functions
used for training.

\myparagraph{Evaluation metrics.}  We use two types of aggregate metrics in our evaluation to assess
both the quality of the minima that were found, as well as the number of iterations an optimizer needed in order to reach the
minimum.

We rely on \emph{performance profiles}~\cite{strogin2000,dolan2002benchmarking,ali2005numerical,beiranvand2017best} to compare \Optimus to the baseline methods. %
Let $\mathcal{P}$ be the set of test problems and $\mathcal{S}$ is the set of optimizers tested.
To define a performance profile one first introduces a performance measure 
$m_{p,s} = \frac{\hat{f}_{p,s} - f^*_{p}}{f^w_{p} - f^*_{p}}$, where $\hat{f}_{p,s}$ is best solution of method $s$ for problem $p$, $f^w_{p}$ is the worst solution out of all methods, and $f^*_{p}$ is the global minimum. The measure $m_{p,s}$ is useful because if allows to compare performance of the optimizer across a set of different optimization tasks. 
The following ratio then compares each optimizer with the best performing one on the problem $p$: $r_{p,s} = \frac{m_{p,s}}{\min\{m_{p,s} : s \in \mathcal{S}\}}$, where the best solver for each problem has ratio $r_{p,s} = 1$. 
Given a threshold $t$ for each optimizer $s$ we can now compute the percentage of problems $\rho_s(t)$ for which the ratio $r_{p,s} \leq t$:
\begin{equation}
    \rho_s(t) = \frac{1}{|\mathcal{P}|}\text{size}\{p\in \mathcal{P}: r_{p,s} \leq t\}.
\end{equation}

\begin{figure}[tb]
  \centering
  \includegraphics[width=0.95\linewidth]{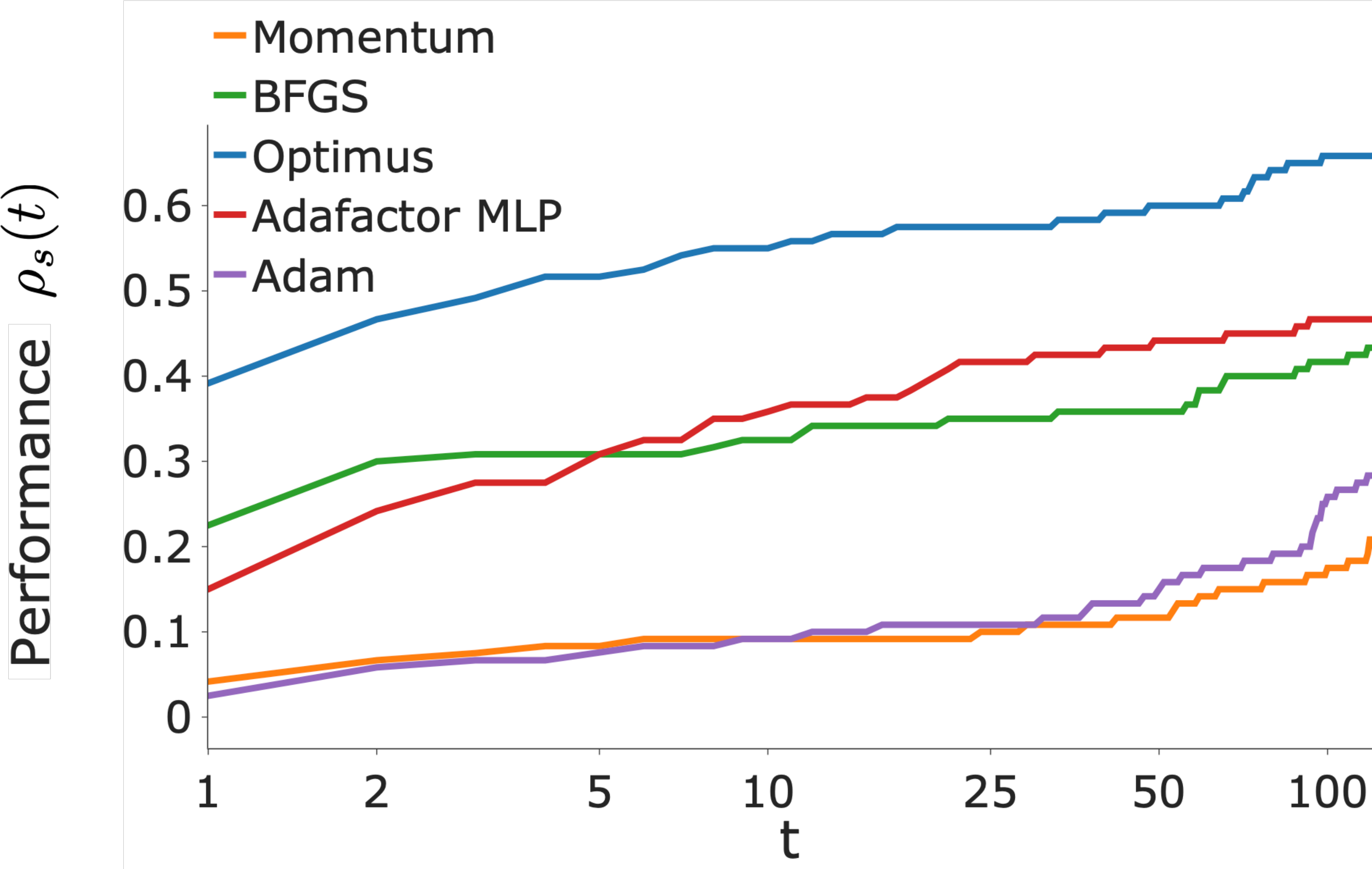}
  \vspace{-2mm}
  \caption{Comparison of optimizers using the performance profile metric from
    \cite{beiranvand2017best} that incorporates relative distance from the global minimum and
    relative performance of each algorithm with respect to the best algorithm (higher values on the
    y-axis mean better performance).}
  \label{fig:opt_results_classical_perf}
  \vspace{-4mm}
\end{figure}

Thus, the performance profile $\rho_s(t)$ is the proportion of problems a method's performance ratio $r_{p,s}$ is within a factor of $t$ of the best performance ratio. Hence, $\rho_s(1)$ represents the percentage of tasks for which optimizer $s$ has best performance (lowest function value) out of the tested methods.
We compare the performance profiles of \Optimus with our baselines in \cref{fig:opt_results_classical_perf}.

Finally, we also measure the relative number of iterations \Optimus{} needed to reach the value of
the minima found by another baseline optimizer. We use Adam, BFGS and \AdafactorMLP as baselines for
such comparison.

\begin{figure}[tb]
  \centering
  \includegraphics[width=0.98\linewidth]{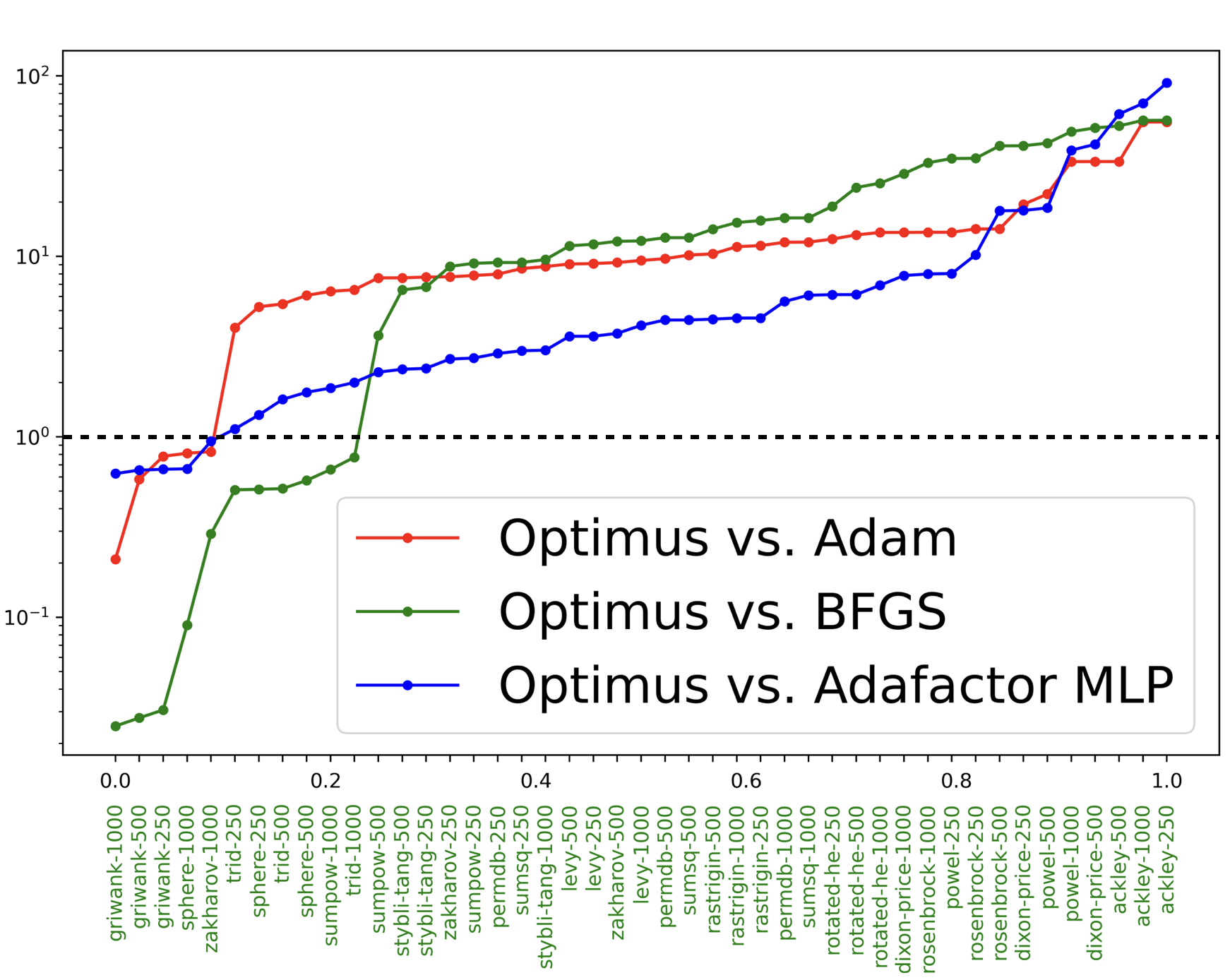}
  \caption{Comparison of \Optimus{} with Adam, BFGS and Adafactor MLP in terms of relative number of
    iterations required to reach a pre-defined minimum of the objective function.
    Black dotted line corresponds to break-even point, all the dots above this line correspond to
    optimization problems where
    \Optimus required fewer iterations than corresponding baseline. The units of the x-axis are
    percentiles of the total number of tasks. For illustration, we mark the points on the x-axis
    with the name and dimensionality of the objective function
    using the ordering of the \Optimus{} vs. BFGS comparison (as indicated by the green color).}
  \label{fig:opt_results_classical_rel_iters}
  \vspace{-2mm}
\end{figure}

\begin{figure}[tb]
  \centering
  \includegraphics[width=0.95\linewidth]{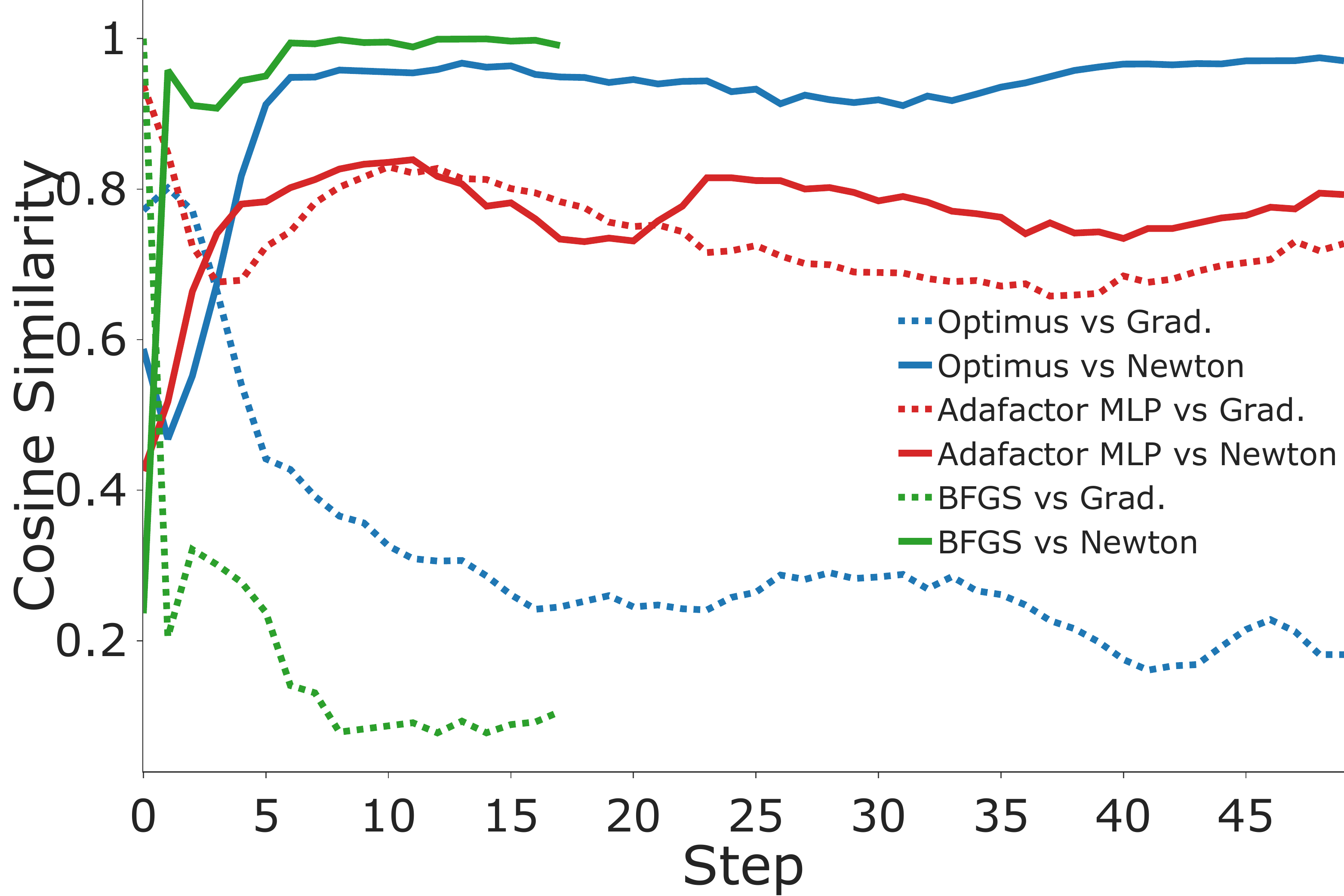}
  \vspace{-2mm}
  \caption{Mean similarity between Optimus update step and gradient and Newton direction on the 2d Rosenbrock function, averaged over 64 trajectories.}
  \vspace{-2mm}
  \label{fig:rosenbrock_10_100d}
\end{figure}

\myparagraph{Results.}  We observe that \Optimus{} typically converges to a lower objective value
compared to other optimizers (see \cref{fig:opt_results_examples} and supplementary material). We use
the performance profile metric \cite{beiranvand2017best} to aggregate these results across functions.
Note that it is not meaningful to directly average the per-function minima since
each function is scaled differently and so values of minima are not directly comparable. The
performance profile metric tackles this issue by relating minima for each function to its
global minima, which is known for all objective functions in our benchmark.
The results are shown \cref{fig:opt_results_classical_perf}. We observe that performance of \Optimus is higher
than other optimizers across all values of performance threshold indicating that on average \Optimus
gets closer to global minimum of each function compared to other optimizers.

In \cref{fig:opt_results_classical_rel_iters} we show relative number of iterations \Optimus needs
to reach the same objective value as Adam, BFGS and \AdafactorMLP optimizers. The target objective
value in this evaluation is defined as average objective value achieved by the corresponding
baseline after $100$ iterations. 
The y-axis in \cref{fig:opt_results_classical_rel_iters} corresponds to ratio between number of
iterations required by a baseline and number of iterations required by \Optimus{}. Values on y-axis
larger than $1$ indicate that \Optimus{} required fewer iterations. The x-axis in
\cref{fig:opt_results_classical_rel_iters} is a percentile of the tasks in the benchmark. A
particular point on the plot then tells us what percentage of the benchmark has a ratio between
number of iterations greater or equal than a value on y axis at that point.

For example, we observe that for about $50\%$ of the tasks \Optimus requires about $10\times$ fewer
iterations than Adam and BFGS and about $5\times$ fewer iterations than \AdafactorMLP.
In \cref{fig:opt_results_classical_rel_iters} we label the x-axis with names of the objective
functions corresponding to each point on the curve comparing \Optimus to BFGS. We observe that 
BFGS excels on simple convex objective functions
such as ``Sphere'' 
or their noisy versions such as
``Griewank'',
where it can quickly converge to global minimum. However BFGS 
fails on
more complex functions with multiple local minima such as ``Ackley'' or functions where minima is
located in a flat elongated valley such as ``Rosenbrock'' or ``Dixon-Price''\footnote{Please
  see supplementary material for the description of the objective functions.}.

\begin{figure*}[tb]
  \centering
  \includegraphics[width=0.32\linewidth]{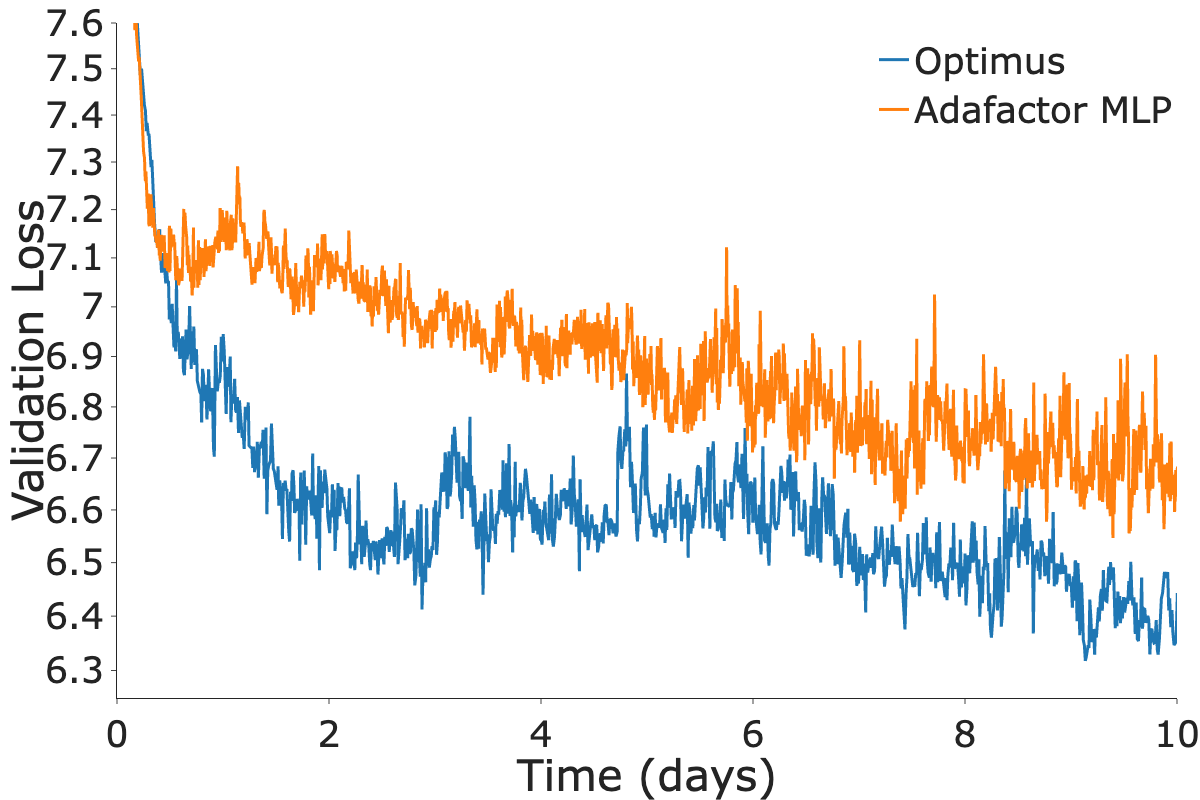}
  \includegraphics[width=0.32\linewidth]{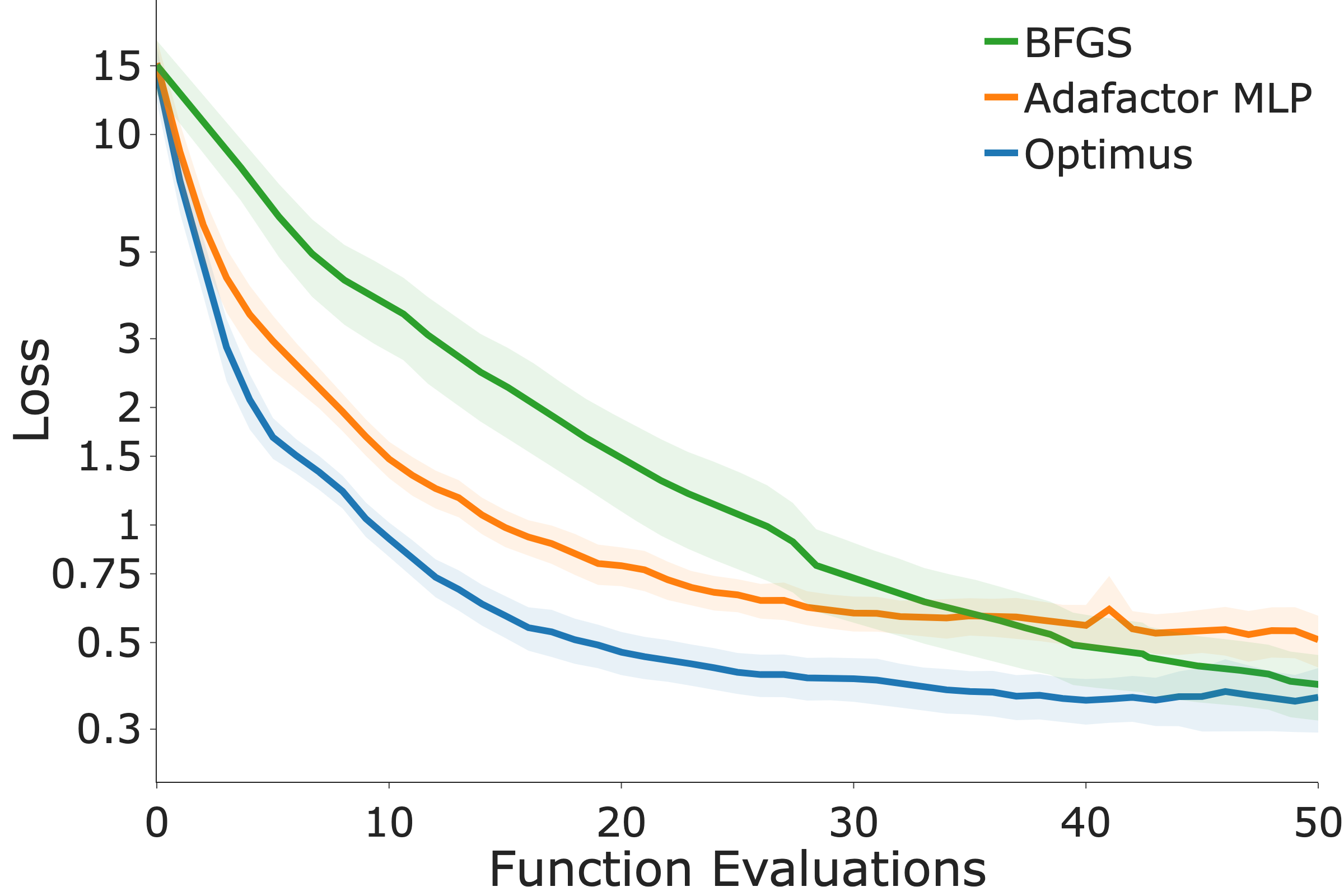}
  \includegraphics[width=0.32\linewidth]{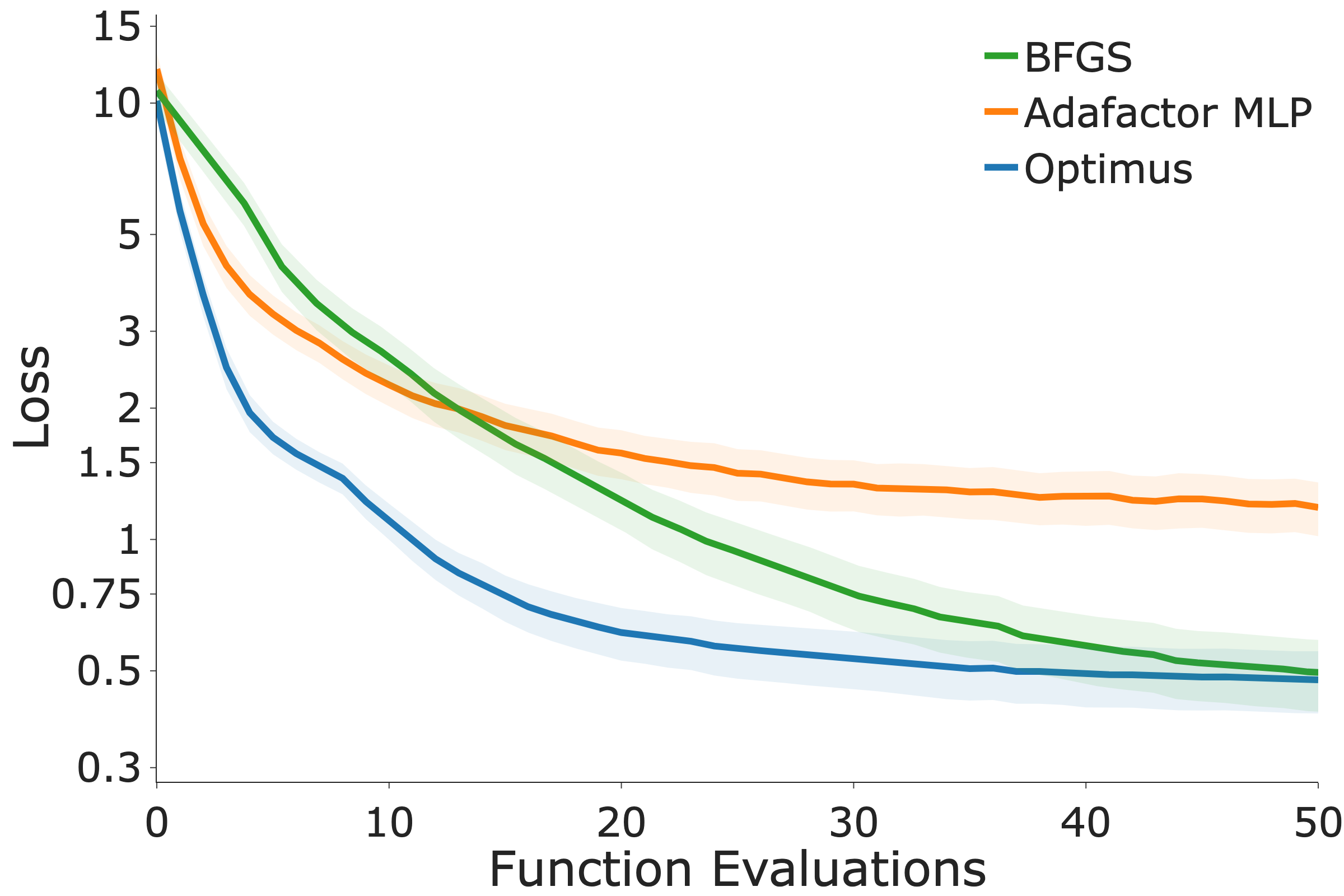}
  \caption{Left: comparison of validation loss during training of Optimus and Adafactor MLP. Note how Optimus converges to lower validation loss much faster than the Adafactor MLP model. Mid: comparison of loss curves during optimizing on ``in domain'' examples in \cref{tab:rl_vs_lopt}. Right: loss curves during optimizing of ``out of domain'' examples in \cref{tab:rl_vs_lopt}. Note how Optimus generalizes better than Adafactor MLP on out of domain data and minimizes the loss faster than BFGS. Shaded area denotes 95\% confidence interval.}
  \label{fig:adafac_comparisons}
\end{figure*}

\begin{table*}[bt]
\begin{center}
\scalebox{0.95}{
\begin{tabular}{l|c|c|c}
\textbf{Method} & \textbf{MPJPE-G (in domain)}  &  \textbf{MPJPE-G (out of domain)} & \textbf{\# Func. Evals} \\
  \hline
  DiffPhy + BFGS ~\cite{gartner2022diffphy} & $38.3$ & $24.7$ & $71$ \\
  $\quad$ + AdaFactor MLP~\cite{metz2022practical} & $33.0$ & $29.3$ & $50$ \\
  $\quad$ + Optimus (Ours) & $24.0$ & $25.0$ & $50$ \\ %
\end{tabular}
}
\end{center}
\vspace{-4mm}
\caption{Comparison of different approaches to trajectory optimization on motion capture data from our Human3.6M validation set.}
\vspace{0mm}
\label{tab:rl_vs_lopt}
\end{table*}

\myparagraph{Analysis of update step direction.}  To further highlight differences between
\Optimus~and \AdafactorMLP~ we plot the absolute cosine similarity of their step along the steepest
descent direction given by $-\nabla\V x$, and the Newton direction given by
$-\M H_{\V x}^{-1}\nabla\V x$, where $\M H_{\V x}^{-1}$ is the inverse Hessian at point $\V x$. The
results are shown in \cref{fig:rosenbrock_10_100d} for optimization of the 2d Rosenbrock
function. For clarity we also include the same similarity plot for BFGS. As expected, the direction
of BFGS step becomes similar to Newton after a few iterations since the preconditioner in BFGS
approximates the inverse Hessian. Note that overall the direction of the \Optimus~step is much
closer to Newton compared to \AdafactorMLP~ and much less similar to steepest descent. This supports
the intuition that \Optimus{}'s design extends the learned optimizer with a preconditioner similar
to BFGS.

\subsection{Physics-Based Motion Reconstruction}
\label{sec:physics_based_hpe}
\begin{table*}[tbhp]
\begin{center}
\scalebox{1}{
\begin{tabular}{l|c|c|c|c|c|c}
\textbf{Model} & \textbf{MPJPE-G} & \textbf{MPJPE} & \textbf{MPJPE-PA} & \textbf{MPJPE-2d} & \textbf{TV} & \textbf{Foot skate} \\
\hline
VIBE~\cite{kocabas20cvpr} & 207.7 & 68.6 & 43.6 & 16.4 & 0.32 & 27.4 \\
PhysCap~\cite{PhysCapTOG2020} & - & 97.4 & 65.1 & - & - & -  \\
SimPoE~\cite{yuan2021simpoe} & - & \textbf{56.7} & \textbf{41.6} & - & - & -  \\
Shimada et al.~\cite{Shimada2021NeuralM3} & - & 76.5 & 58.2 & - & - & -  \\
Xie et al.~\cite{Xie_2021_ICCV} & - & 68.1 & - & - & - & -  \\

DiffPhy~\cite{gartner2022diffphy} & 139.1 & 82.1 & 55.9 & 13.2 & 0.21 & 7.2  \\
\hline

Optimus (Ours) & \textbf{138.6} & 82.8 & 57.0 & 13.2 & \textbf{0.20} & \textbf{6.5} \\ %

\end{tabular}
}
\end{center}
\vspace{-4mm}
\caption{Quantitative results on Human3.6M~\cite{h36mpami} comparing our model to prior methods.}
\label{tab:quantitative}
\end{table*}
\begin{table*}[htbp]
\begin{center}
\scalebox{1}{
\begin{tabular}{l|c|c|c|c|c|c}
\textbf{Model} & \textbf{MPJPE-G} & \textbf{MPJPE} & \textbf{MPJPE-PA} & \textbf{MPJPE-2d} & \textbf{TV} & \textbf{Foot skate} \\
\hline
DiffPhy~\cite{gartner2022diffphy} & 150.2 & 105.5 & 66.0 & 12.1 & 0.44 & 19.6 \\
Optimus & 149.8 & 104.4 & 66.4 & 12.1 & 0.45 & 21.5  \\
\end{tabular}
}
\end{center}
\vspace{-2mm}
\caption{Quantitative results on generalizing to a new dataset. Optimus was trained Human3.6M~\cite{h36mpami} and is here evaluated on a subset of the dance motion dataset AIST~\cite{aist-dance-db}.}
\vspace{0mm}
\label{tab:aist}

\end{table*}
\begin{table*}[!th]
\begin{center}
\scalebox{1}{
\begin{tabular}{l|c|c|r|r}
\textbf{Variant} & \textbf{MPJPE-G} & \textbf{Loss} & \textbf{\# Params} & \textbf{Runtime (ms)}  \\
  \hline
  Optimus & 33.1 & 0.534 & 832,526 & 78.9 \\
  Optimus, no state & 36.5 & 0.675  & 832,526 & 115.1 \\
  Optimus, no structure & 37.3 & 0.721  & 832,139 & 10.7 \\
  Adafactor MLP~\cite{metz2022practical} 512x4 & 40.0 & 0.723  & 811,531 & 3.7 \\
  Adafactor MLP~\cite{metz2022practical} 128x4  & 40.6 & 0.811  & 22,411 & 2.6 \\
  \hline
  Optimus & 33.1 & 0.534  & 832,526 & 78.9 \\
  Optimus, 50 iterations  & 28.5 & 0.383 & 832,526 & 78.9 \\
  Optimus + BasinHopping~\cite{wales1997basin} & 25.6 & 0.352  & 832,526 & 78.9 \\
\end{tabular}
}
\end{center}
\vspace{-2mm}
\caption{Top: ablation of model components on motion capture from our Human3.6M~\cite{h36mpami} validation set. Bottom: comparison of different test time operating modes.}
\vspace{-2mm}
\label{tab:model_ablation}
\end{table*}

\begin{figure}[tb]
  \centering
  \includegraphics[width=1.0\linewidth]{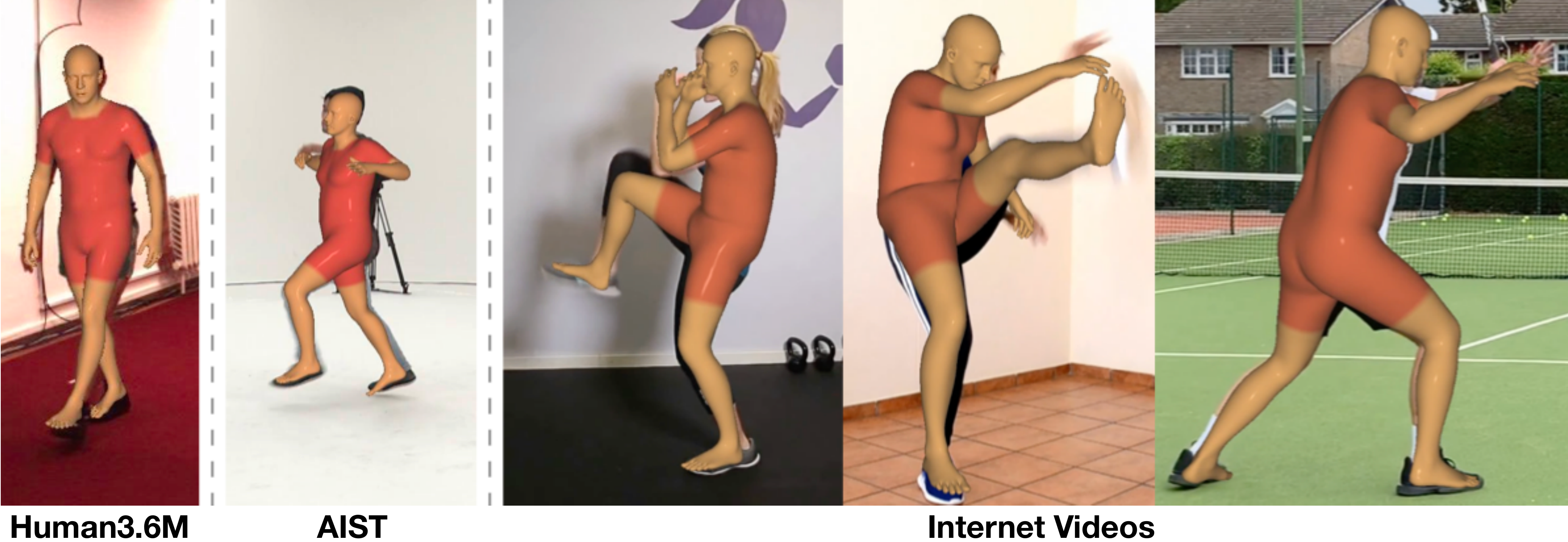}
  \vspace{-4mm}
  \caption{Qualitative examples of video reconstructions by \Optimus{}. We train \Optimus{} on the Human3.6M~\cite{h36mpami} dataset (left). We then evaluate its performance on the dancing sequence from AIST~\cite{aist-dance-db} (see \cref{tab:aist}) and qualitatively verify that \Optimus{} is applicable to in-the-wild internet videos (rightmost images).}
  \label{fig:qualitative_examples}
  \vspace{-2mm}
\end{figure}

In principle, learned optimization can be applied to any problem that is solvable by means of local descent, and thus to any physics-based reconstruction that formulates human motion reconstruction as loss minimization
\cite{li2019cvpr,zanfir2020weakly,gartner2022diffphy,Xie_2021_ICCV}. 
In this paper, we build on the \DiffPhy{} approach of \cite{gartner2022diffphy}, who define a differentiable loss function for physics-based reconstruction.
DiffPhy relies on the differentiable implementation of rigid body dynamics in \cite{heiden2021neuralsim} and shape-specific body model based on \cite{xu2020ghum} to define a loss function that measures similarity between simulated motion and observations.
The observations are either a set of 3d keyframe poses, when the goal is to reproduce articulated 3d motion in physical simulation, or a sequence of image measurements such as 2d image keypoints or estimated 3d poses in each frame.
The physical motion in DiffPhy is parameterized via a control trajectory, which is given by a sequence of quaternions defining a target rotation of each body joint over time.
The control trajectory implicitly defines the torques applied to each body joint via PD-control. Please refer to \cite{gartner2022diffphy} for a more extensive explanation of the loss function.
Hence, optimization, in this case, aims to infer the control trajectory that re-creates an articulated motion in physical simulation in a way that is close to the given observations and consistent with constraints (\eg lack of foot-skate and non-intersection with respect to a ground plane).

\myparagraph{Datasets.} In our human motion experiments, we use the popular Human3.6M articulated pose estimation benchmark~\cite{h36mpami}. We follow the protocol introduced in \cite{PhysCapTOG2020} to compare to related work for video-based experiments\footnote{The protocol excludes motions such as ``sitting'' and ``eating'' which require modeling human-object interactions.}. For experiments with motion capture inputs, we rely the same subset of $20$ validation sequences from Human3.6M as used in \cite{gartner2022diffphy}. We then additionally evaluate on the dancing sequences from the AIST dataset \cite{aist-dance-db} used in \cite{gartner2022diffphy} to further evaluate the generalization of our approach across qualitatively different motion types.

\myparagraph{Evaluation metrics.} We report the standard 2d and 3d human pose metrics as well as
physics-based metrics. The mean per-joint 3d position error (MPJPE-G), the per-frame
translation-aligned error (MPJPE), and the per-frame Procrustes-aligned error (MPJPE-PA) are reported in
millimeters. In addition, we measure the 2d keypoint error of the reconstruction (MPJPE-2d) in
pixels. Finally, we measure the amount of motion jitter as the total variation in 3d joint
acceleration (TV) defined as
$\frac{1}{T}\sum_{t\in T}\sum_{k \in K} |\ddot{x}^k_{t+1} -\ddot{x}^k_{t}|$, where $\ddot{x}^k_{t}$
is the 3d acceleration of joint $k$ at time $t$, as well as the percentage of frames exhibiting
\emph{foot skate} as defined by prior work~\cite{gartner2022diffphy}.

\myparagraph{Model training.} We train \Optimus{} and \AdafactorMLP{} on the task of minimizing the
DiffPhy loss, which measures similarity between the simulated motion and 3d poses estimated based on \cite{zanfir2020neural}. 
Following an initial grid-search to determine the best learning rate for each model, we train these
for 10 days using 500 CPUs to generate training batches of optimization rollouts (128 roll-outs of length 50 per batch). 
We show loss vs. training time curves on the validation set in \cref{fig:adafac_comparisons} (left).
Note that \Optimus{} generally converges much faster and to a lower loss value compared to competitors. For example, after $48$
hours of training, \Optimus{} has essentially converged to a loss of $6.49$ whereas \AdafactorMLP{} requires nearly $240$ hours to reach a loss value of $6.55$.

\myparagraph{Results.}
As our first experiment in \cref{tab:rl_vs_lopt} we compare the performance of \Optimus to \AdafactorMLP and BFGS using motion capture (mocap) data as input.
In \cref{fig:adafac_comparisons} (middle and right) we additionally visualize how quickly the
loss is reduced by the optimizer at each iteration. Note that iterations of BFGS and of the
learned optimizers are not comparable in terms of computational resources because BFGS might
evaluate the objective function more than once during line search. To compensate for different iteration costs, in \cref{fig:adafac_comparisons} we
plot the number of objective function evaluations instead of the optimization iteration along the x-axis.
In this experiment we train all models on walking sequences from Human3.6M dataset \footnote{We use ``Walking'' and ``WalkTogether'' activities.}. We refer to walking sequences as ``in domain'' in \cref{fig:adafac_comparisons} and \cref{tab:rl_vs_lopt}. We then assess performance on a validation set of ``in domain'' and ``out of domain'' sequences corresponding to
motions other than walking.
Note that \Optimus{} improves over other approaches in the
``in domain'' setting and reaches the loss comparable to BFGS at roughly half of loss function
evaluations (\cf \cref{fig:adafac_comparisons} (middle)). In the ``out of domain'' setting
\Optimus{} reaches nearly the same accuracy compared to BFGS ($24.7$ vs $25$mm.) but again
converges much faster. In contrast \AdafactorMLP{} does not improve over BFGS on the ``out of domain''
motions and converges to a higher loss (\cf \cref{fig:adafac_comparisons} (right)). 

In the second experiment, we evaluate performance of our best model \mbox{DiffPhy+\Optimus{}} for video-based human motion reconstruction. 
Note that \Optimus{} performs well on the dancing sequences from AIST even though it has been trained only on walking data from Human3.6M ($150.2$ for DiffPhy vs. $149.8$~mm MPJPE-G for \Optimus) and that \Optimus{} is able to
handle mocap and video inputs without retraining. We observe similar results on Human3.6M, where \Optimus{} performs on par with BFGS ($138.6$ vs. $139.1$ MPJPE-G) while requiring
roughly half as many function evaluations ($88$ for \DiffPhy{}~vs.~$40$ for \Optimus{}).
We show a few qualitative
examples for \Optimus{} results on Human3.6M, AIST and internet videos in \cref{fig:qualitative_examples}.

\myparagraph{Ablation experiments.}
We now evaluate how different design choices affect the model performance. Due to high computational
load, we train each model for up to $48$ hours.
The results are shown in 
\cref{tab:model_ablation}. We refer to a version of \Optimus~that does not incrementally
update the matrix $\M B^{k}$ in \cref{eq:optimus} and instead predicts it from scratch at each
iteration as ``\Optimus{}, no state'', and a version that uses a transformer to directly predict
$\Delta \V x^{k}$ from $\V s^k$ in \cref{eq:optimus_update} as ``\Optimus, no structure''. 
Our full \Optimus{} improves over \AdafactorMLP{} by $17.2\%$ and over simpler ``\Optimus{}, no
structure'' by $11.3\%$ while having nearly the same number of parameters. Note that simply
adding more parameters to \AdafactorMLP{} barely improves results ($40.6$ vs. $40.0$mm MPJPE-G).
\Optimus{} makes $30.3$ function evaluations before optimization is terminated by the
stopping criterion. In \cref{tab:model_ablation} (bottom three lines) we evaluate the effect of running
\Optimus{} for a fixed number of $50$ function evaluation without stopping criterion and the effect of using even
more costly BasinHopping~\cite{wales1997basin} optimization that adds random perturbations after a
fixed number of iterations (to improve global exploration) requiring $82.3$ function evaluations on average. We observe that running
\Optimus{} for longer leads to considerably improved results, at higher computational load
($33.1$ for \Optimus{} vs. $25.6$~mm MPJPE-G for \Optimus{}+BasinHopping).

\section{Conclusion}
We have introduced a learned optimizer, {\bf \Optimus{}}, based on an expressive architecture that can capture complex dependency updates in parameter space. Furthermore, we have demonstrated the effectiveness of \Optimus{} for the real-world task of physics-based articulated 3D motion reconstruction as well as on a benchmark of classical optimization problems.
While \Optimus{}'s expressive architecture outperforms simpler methods such as \AdafactorMLP{}, the expressiveness comes at an increased computational cost.
As a result, \Optimus{} is best suited for tasks where the loss function dominates the computational complexity of optimization (\eg, physics-based reconstruction) but might be less suited for applications where the computation of the loss function is fast (\eg~training neural networks). 
In future work, we hope to address this limitation by learning factorizations of the estimated prediction matrix.\\
\myparagraph{Ethical Considerations.} We aim to improve the realism and quality of human pose reconstruction by including physical constraints. By amortizing the computation through learning from past instances, we hope to reduce the long-term computational demand of these methods. We believe that our physical model's level of detail (e.g. lack of photorealistic appearance) limits its applications in adverse tasks such as person identification or deepfakes. Furthermore, the model is inclusive in supporting a variety of body shapes and sizes, and their underlying physics.

\clearpage
{\small
\bibliographystyle{ieee_fullname}
\bibliography{biblio}
}

\clearpage

\newcommand{\citep}[1]{\cite{#1}}
\newcommand{\citet}[1]{\cite{#1}}

\appendix
\section*{Appendix}

In this supplementary material, we include additional results and visualizations (\cref{sec:sm_additional_results}), describe details of our implementation of \Optimus{} (\cref{sec:sm_implementation_details}), list the input features  (\cref{sec:sm_features}), 
summarize the hyperparameters of our method (\cref{sec:sm_hyperparameters}) and include the details of the datasets used in the paper (\cref{sec:sm_datasets}).

\section{Additional visualizations}\label{sec:sm_additional_results}
The following section provides additional visualizations of the experiments in the main paper.

\myparagraph{Results on standard evaluation functions.} We evaluate \Optimus on a benchmark of 15 classical optimization test functions. \Cref{fig:classical_function_results} and \cref{tab:full_classical_results} provides detailed per-function results. These results were then aggregated to calculate the performance profile (fig. 3 in the main paper).

\myparagraph{Comparison of optimizer efficiency.} \Cref{fig:rosenbrock_fevals_vs_loss} shows the
efficiency of \Optimus{} on $N$-dimensional Rosenbrock functions and compares it to other
optimization algorithms. Note that \Optimus{} achieves better loss values while using significantly less compute than BFGS and \AdafactorMLP{}. For $N=1000$, \Optimus{} converges after $\myapprox 60$ function evaluations achieving a loss value $\myapprox 5$ while BFGS achieves a loss of $\myapprox 10^6$ using $\myapprox 240$ function evaluations. The plot was generated using the same Rosenbrock optimization trajectories as presented in the main paper.

\begin{figure}[bht]
  \centering
  \includegraphics[width=1\linewidth]{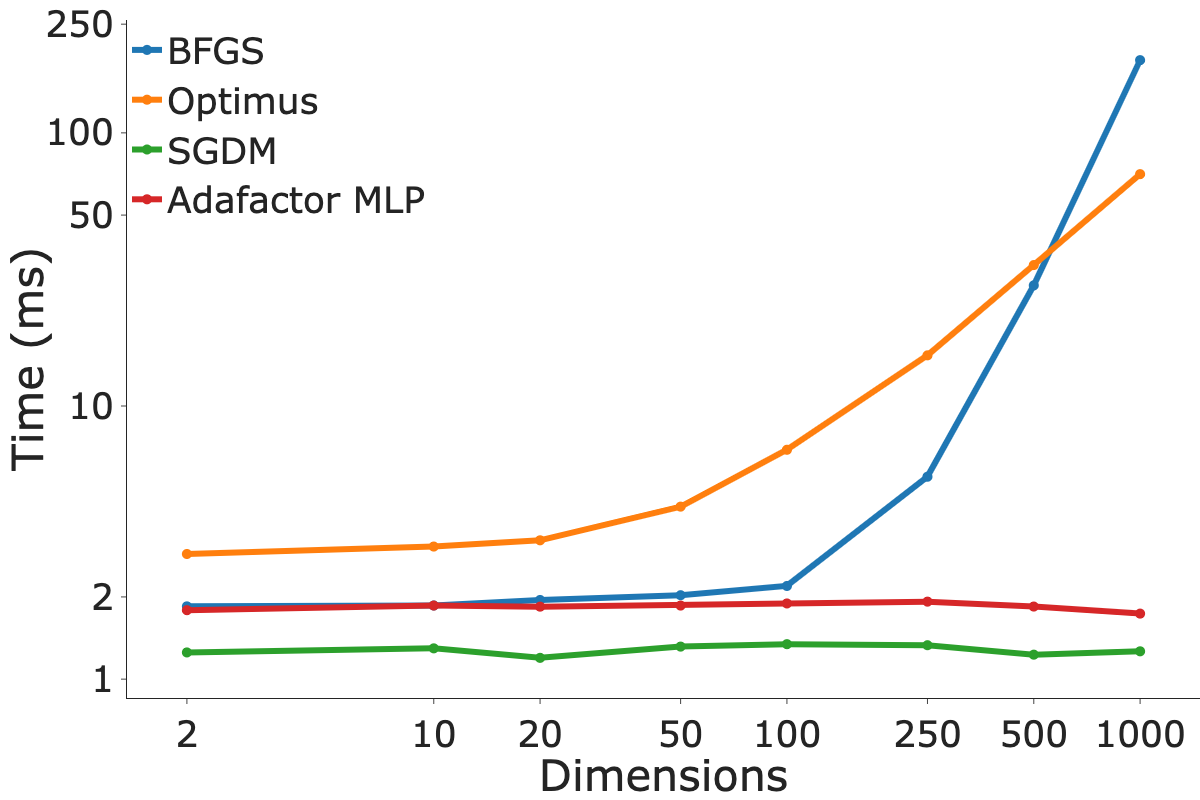}
  \caption{Time required to compute the optimizer update step as a function of dimensionality of
    the objective function.}
  \label{fig:rosenbrock_timings}
  \vspace{-2mm}
\end{figure}

\myparagraph{Runtime Comparison.}  In \cref{fig:rosenbrock_timings} we plot the runtime of the optimization step 
for \Optimus{} and other optimization methods on the task of finding minima of 
$N$-dimensional Rosenbrock functions while varying $N$ between $2$ and $1000$. In this comparison we use a widely
adopted implementation of BFGS from the SciPy~\cite{2020SciPy-NMeth} package.  Both \Optimus{}
and BFGS~\cite{Flet87} have $\mathcal{O}(N^2)$ time complexity. Interestingly for large $N$
\Optimus{} optimization step is faster than BFGS, which is likely to be due to its more efficient
Jax~\cite{jax2018github} implementation.

\myparagraph{Additional Optimization Trajectories.} \Cref{fig:more_rosenbrock_trajectories} presents
additional visualizations of optimization trajectories on $N$-dimensional Rosenbrock functions. 
Note how \Optimus{} tends to vary
its step size throughout the trajectories while \AdafactorMLP{} tends to monotonically decrease its
step size. We have observed that \AdafactorMLP{} generally tends to learn a learning rate schedule
based on the current iteration (a feature to the networks) and decrease its step size monotonically.

\section{Implementation details}\label{sec:sm_implementation_details}

\begin{figure}[tb]
  \centering
   \includegraphics[width=1.0\linewidth]{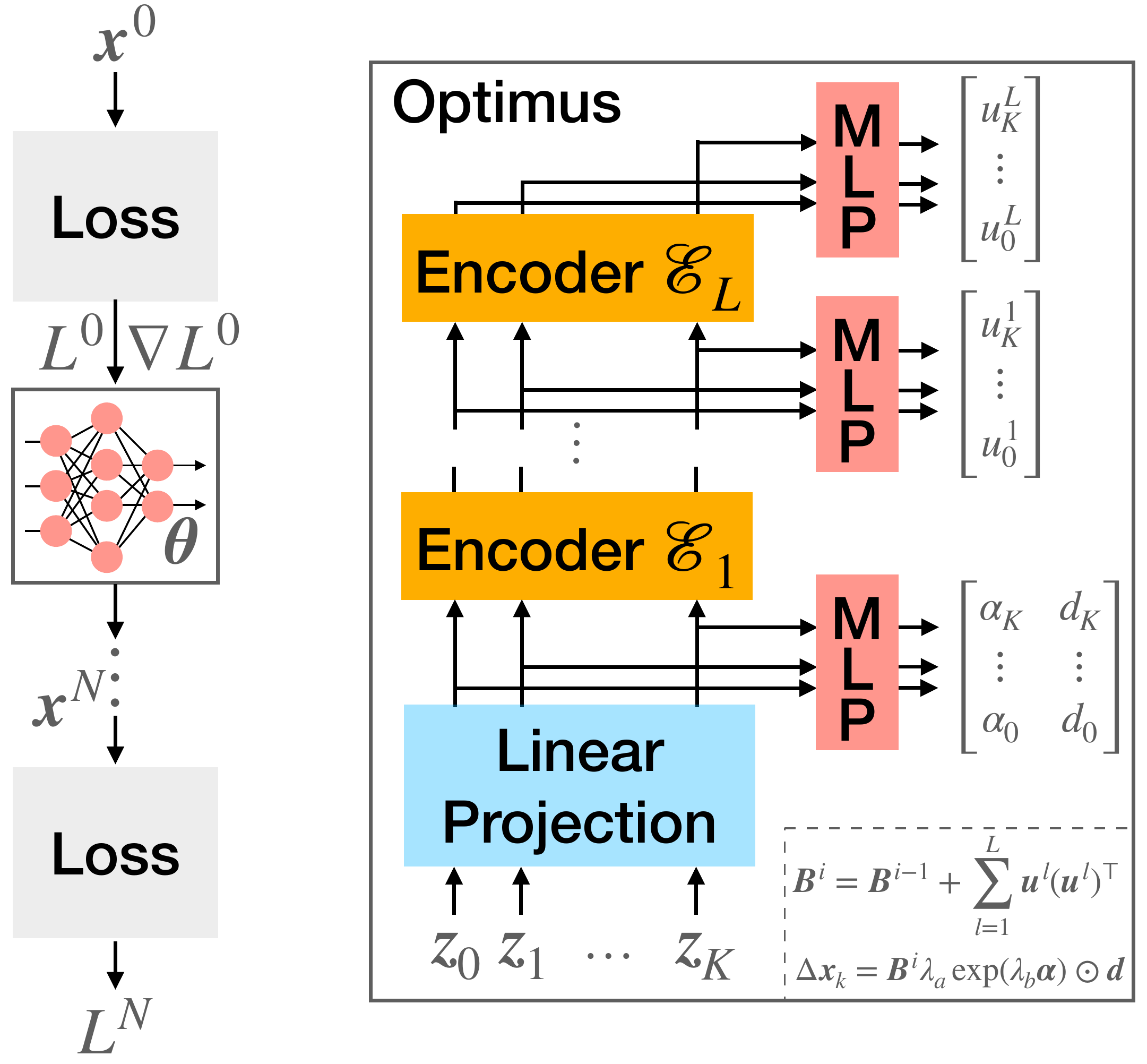}
   \vspace{-2mm}
  \caption{Left: schematic overview of applying our Transformer-based learned optimizer, \emph{Optimus}, to iteratively minimize the loss $L$. Right: architecture of Optimus consisting of $L$ stacked Transformer encoders that predict the parameter update $\Delta \V x_k$ given the feature vector $\V z_k$ consisting of the associated gradient $\frac{dL}{d\V x_k}$ together with the features from \cite{metz2022practical}.}
   \vspace{-2mm}
  \label{fig:optimus}
\end{figure}

The \Optimus{} algorithm is shown in alg.~\ref{alg:optimus}
and the corresponding neural network architecture is shown in \cref{fig:optimus}.

\myparagraph{Training.} We implement \Optimus{} in Jax~\cite{jax2018github} using
Haiku~\cite{haiku2020github} and the
\emph{learned\_optimization}\footnote{\url{https://github.com/google/learned_optimization}}
framework. When training \Optimus{} on human motion reconstruction task we using a batch size of
$20$. We generate the batches by sampling random windows from the Human3.6M~\citep{h36mpami}
training set, then rolling out the optimization for $50$ steps, and computing the
PES~\cite{pmlr-v139-vicol21a} gradients every $5$ steps using $2$ antithetic samples. We learn the
weights using Adam~\cite{kingma2014adam} with the learning rate of $5\times 10^{-4}$. To stabilize
the training we perform gradient clipping to gradients with a norm greater than $3$.

\begin{figure}[tbp]
  \centering
   \includegraphics[width=1.0\linewidth]{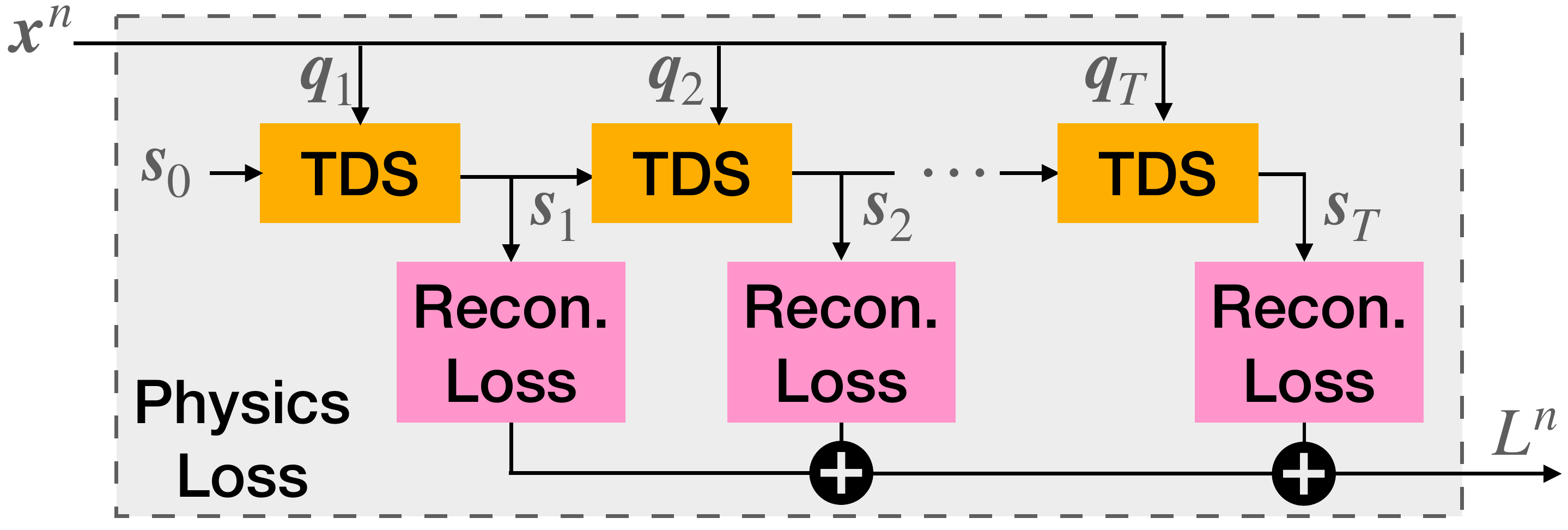}
   \vspace{-2mm}
  \caption{Overview of the physics loss introduced in \cite{gartner2022diffphy} where the optimization variable $\V x$ corresponds to joint torques of a physically simulated human with the goal of minimizing a pose reconstruction loss between the physical character and observed visual evidence.}
  \label{fig:tds_loss}
  \vspace{-4mm}
\end{figure}

\subsection{Distributed Physics Loss}
We use the differentiable physics loss introduced in \citet{gartner2022diffphy} using the Tiny Differentiable Simulator~\citep{heiden2021neuralsim} (TDS) that was implemented in C++. A schematic overview of the loss function is available in \cref{fig:tds_loss}, however, see \cite{gartner2022diffphy} for an in-depth explanation. The gradients of the loss are computed using the automatic differentiation framework CppAD~\cite{cppad}. We wrap the simulation step function as a custom TensorFlow~\cite{tensorflow2015-whitepaper} function to enable easier integration with Jax when training \Optimus{}. We sample rollouts from $400$ distributed servers exposing the loss function using the Courier\footnote{\url{https://github.com/deepmind/launchpad/tree/master/courier}} framework. We do this to overcome the issue of the slow evaluation time of the physics loss.

\section{Input features }\label{sec:sm_features}
In the following we list the features $\V z$ used as input by \Optimus{}. The features are identical to the features used in \cite{metz2022practical} and we list them here to make the paper more self-contained. 

\begin{itemize}
\item the parameter values
\item the 3 momentum values ($m$)
\item the second moment value ($v$)
\item 3 values consisting of momenta normalized by rms gradient norm $-m/\sqrt v$
\item the $(\sqrt{v+\epsilon})^{-1}$
\item 3 AdaFactor normalized gradient values
\item the tiled, AdaFactor row features (3 features)
\item the tiled, AdaFactor column features (3 features)
\item $1.0$ divided by the square root of these previous $6$ AdaFactor features
\item 3 features consisting AdaFactor normalized momentum values
\item 11~features formed by taking the current timestep, $t$, and computing $\mbox{tanh}(t/x)$ where $x \in \{1, 3, 10, 30, 100, 300, 1000, 3000, 10k, 30k, 100k\}$.
\end{itemize}

\section{Hyperparameters and Computional Resources}\label{sec:sm_hyperparameters}
We train \Optimus using a distributed setup on the Google Compute Engine\footnote{\url{https://cloud.google.com/compute}}. When training on the DiffPhy~\citep{gartner2022diffphy} reconstruction task, we distributed the loss function on $400$ vCPU instances. This significant computational cost is what motivated us to find a model which converges faster than prior work (e.g. \AdafactorMLP~\citep{metz2022practical}). We trained approximately $30$ such models throughout experimentation.

Training models on $N$-dimensional Rosenbrock functions was much faster as its loss function evaluates in $\myapprox 5$ms rather than $\myapprox 4$s. We trained those models for $48$ hours using $40$ vCPU instances. In total, we trained approximately $100$ such models.

The primary hyperparameters and the ranges of them that we tested are presented in \cref{tab:hyperparams}. Due to the high computational expense, we could not test all combinations exhaustively. We chose a short truncation length ($5$) on the physics tasks as it allows us to update the network more often which speed up the training convergence. Similarly, having a smaller batch size allowed us to update the model more frequently as gathering training batches was faster.

\begin{table}[b]
     \centering
     \scalebox{1}{
     \begin{tabular}{c|c}
         \textbf{Sequence} & \textbf{Frames}  \\
         \hline
gBR\_sBM\_c06\_d06\_mBR4\_ch06 & 1-120 \\
gBR\_sBM\_c07\_d06\_mBR4\_ch02 & 1-120 \\
gBR\_sBM\_c08\_d05\_mBR1\_ch01 & 1-120 \\
gBR\_sFM\_c03\_d04\_mBR0\_ch01 & 1-120 \\
gJB\_sBM\_c02\_d09\_mJB3\_ch10 & 1-120 \\
gKR\_sBM\_c09\_d30\_mKR5\_ch05 & 1-120 \\
gLH\_sBM\_c04\_d18\_mLH5\_ch07 & 1-120 \\
gLH\_sBM\_c07\_d18\_mLH4\_ch03 & 1-120 \\
gLH\_sBM\_c09\_d17\_mLH1\_ch02 & 1-120 \\
gLH\_sFM\_c03\_d18\_mLH0\_ch15 & 1-120 \\
gLO\_sBM\_c05\_d14\_mLO4\_ch07 & 1-120 \\
gLO\_sBM\_c07\_d15\_mLO4\_ch09 & 1-120 \\
gLO\_sFM\_c02\_d15\_mLO4\_ch21 & 1-120 \\
gMH\_sBM\_c01\_d24\_mMH3\_ch02 & 1-120 \\
gMH\_sBM\_c05\_d24\_mMH4\_ch07 & 1-120
     \end{tabular}
     }
     \caption{Sequences from the dance dataset AIST~\cite{aist-dance-db} used for evaluation.}
 \label{tab:aist_sequences}
\end{table}

\section{Datasets Details}\label{sec:sm_datasets}
We use three sets of data in the paper. Firstly, we evaluate on the established Human3.6M~\citep{h36mpami} dataset, which is recorded in a laboratory setting with the permission of the actors. We list the sequences used in our validation set in \cref{tab:h36m_sequences}. When comparing to prior work we use the protocol established in \citet{PhysCapTOG2020} namely evaluating on sequences \textit{Directions, Discussions, Greeting, Posing, Purchases, Taking Photos, Waiting, Walking, Walking Dog and Walking Together} from subjects S9 and S11. We evaluate the motions using only camera \emph{60457274} and following prior work~\citep{PhysCapTOG2020,Xie_2021_ICCV} we downsample the sequences from $50$ FPS to $25$ FPS.

Next, we use the AIST\footnote{\url{https://aistdancedb.ongaaccel.jp/}}~\cite{aist-dance-db} dataset which features professional dancers performing to copyright-cleared dance music. The sequences we evaluate on are given in \cref{tab:aist_sequences}.
Finally, we provide qualitative examples of our method on ``in-the-wild`` internet videos that were released under creative common licenses. 

\begin{table}[t]
    \centering
    \begin{tabular}{c|c|c|c}
        \textbf{Sequence} & \textbf{Subject} &
        \textbf{Camera Id} & \textbf{Frames}  \\
        \hline
        Phoning & S11 & 55011271 & 400-599 \\
        Posing\_1 & S11 & 58860488 & 400-599 \\
        Purchases & S11 & 60457274 & 400-599 \\
        SittingDown\_1 & S11 & 54138969 & 400-599 \\
        Smoking\_1 & S11 & 54138969 & 400-599 \\
        TakingPhoto\_1 & S11 & 54138969 & 400-599 \\
        Waiting\_1 & S11 & 58860488 & 400-599 \\
        WalkDog & S11 & 58860488 & 400-599 \\
        WalkTogether & S11 & 55011271 & 400-599 \\
        Walking\_1 & S11 & 55011271 & 400-599 \\
        Greeting\_1 & S9 & 54138969 & 400-599 \\
        Phoning\_1 & S9 & 54138969 & 400-599 \\
        Purchases & S9 & 60457274 & 400-599 \\
        SittingDown & S9 & 55011271 & 400-599 \\
        Smoking & S9 & 60457274 & 400-599 \\
        TakingPhoto & S9 & 60457274 & 400-599 \\
        Waiting & S9 & 60457274 & 400-599 \\
        WalkDog\_1 & S9 & 54138969 & 400-599 \\
        WalkTogether\_1 & S9 & 55011271 & 400-599 \\
        Walking & S9 & 58860488 & 400-599 \\
    \end{tabular}
    \caption{Sequences from the Human3.6M~\cite{h36mpami} pose dataset used in our ablations and experiments with reinforcement learning agents.}
    \vspace{-2mm}
    \label{tab:h36m_sequences}
\end{table}
\begin{algorithm*}[!h]
\begin{algorithmic}
\State Initial guess $\xx^{0}$
\State $\boldsymbol{B}^0 \gets \boldsymbol{I}$
\State $k \gets 0$
\Repeat 
    \State $\{\V \alpha, \V d, \Delta \boldsymbol{B}^{k}\} \gets \text{Optimus}(\V z^{k})$
    \State $\boldsymbol{B}^{k} \gets \boldsymbol{B}^{k-1} + \Delta \boldsymbol{B}^{k}$ \Comment{Update learned preconditioning matrix.}
    \State $\bs^k \gets \M B^k [\lambda_a\exp(\lambda_b \balpha) \odot \dd]$
    \State $\xx^k \gets \xx^{k-1} + \bs^k$
    \State $k \gets k + 1$
\Until{$k > \text{MAX\_ITERS or} f(\V x^k) > \frac{1}{N}\sum_{i=1}^{N}\beta f(\V x^{k-i}) + \epsilon$} 

\end{algorithmic}
\caption{The Optimus algorithm. $\V z^k$ denotes the \AdafactorMLP{} features from \citet{metz2022practical} and $\odot$ denotes elementwise multiplication.
}
\label{alg:optimus}
\end{algorithm*}

\begin{table*}[bt]
    \centering
    \scalebox{1}{
    \begin{tabular}{l|c|c}
         \textbf{Variable} & \textbf{Search grid} & \textbf{Value Used}  \\
         \hline
         Learning Rate & \{0.1, 0.01, 0.001, $5 \times 10^{-4}$, $1 \times 10^{-4}$\} & $5 \times 10^{-4}$ \\
         Model Dimension & \{64, 128, 256, 512\} & 128 \\
         \# Encoders & \{1, 2, 3, 4, 5\} & 3 \\
         $\lambda_a$ & \{0.001, 0.01, 0.1\} & 0.1 \\
         $\lambda_b$ & \{0.001, 0.01, 0.1\} & 0.1 \\
         Batch Size & \{20, 50, 128\} & 20 \\
         PES truncation length & \{5, 10, 20\} & 5 \\
    \end{tabular}
    }
    \caption{An overview of the different hyperparameters tested during designing of \Optimus. \emph{Value Used} refers to the value used on the human pose reconstruction task from video.}
    \label{tab:hyperparams}
\end{table*}
\begin{figure*}[tbhp]
  \centering
  \includegraphics[width=0.8\linewidth]{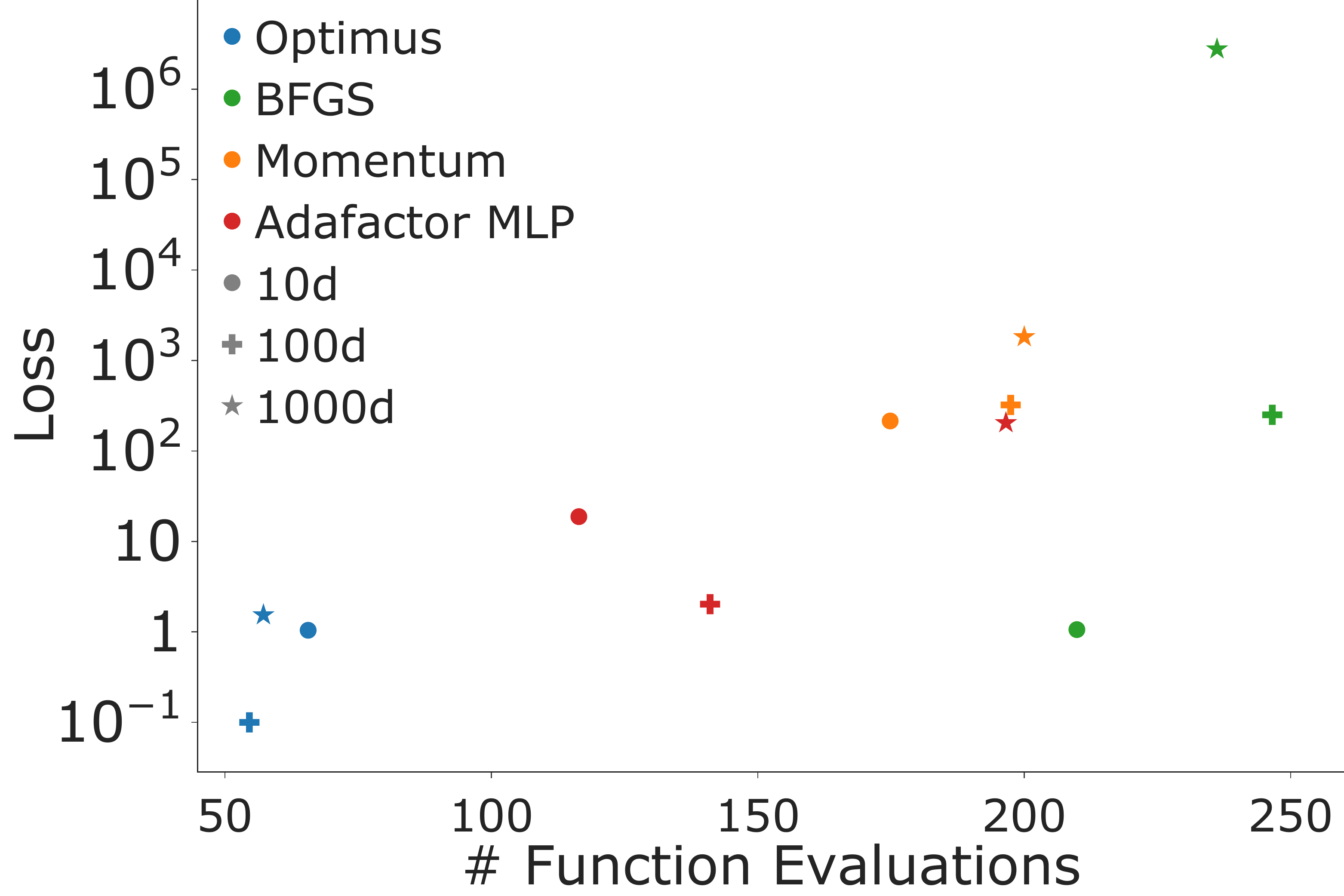}
  \caption{Comparison of loss vs number of function evaluations on the $N$-dimensional Rosenbrock functions for a maximum budget of $200$ steps. Note that BFGS may evaluate the function multiple times per step due to its line search.}
  \label{fig:rosenbrock_fevals_vs_loss}
\end{figure*}

\begin{figure*}[p]
\centering
\begin{tabular}{ccc}

\subfloat[Ackley]{\includegraphics[width = 0.30\linewidth]{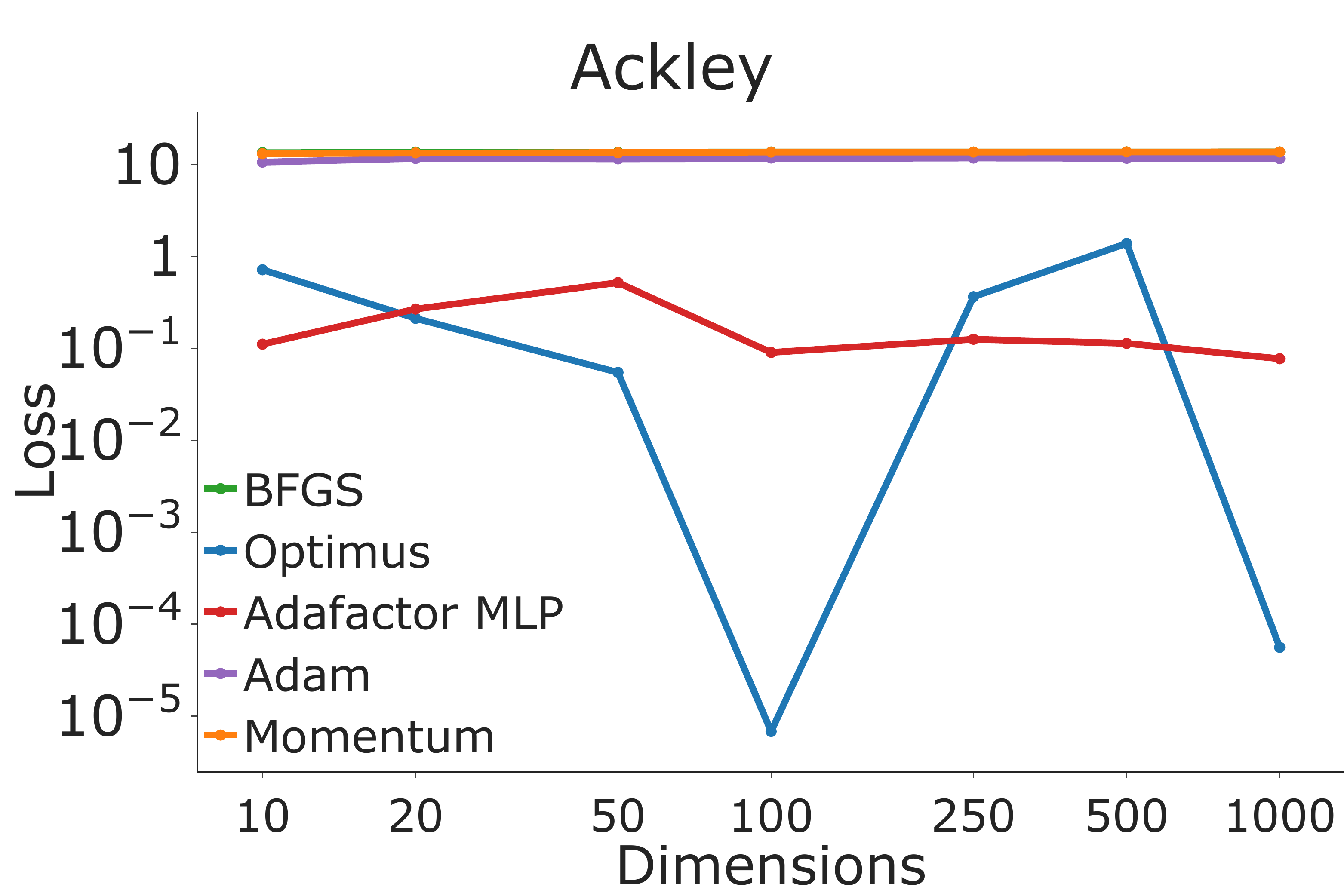}} &
\subfloat[Dixon-Price]{\includegraphics[width = 0.30\linewidth]{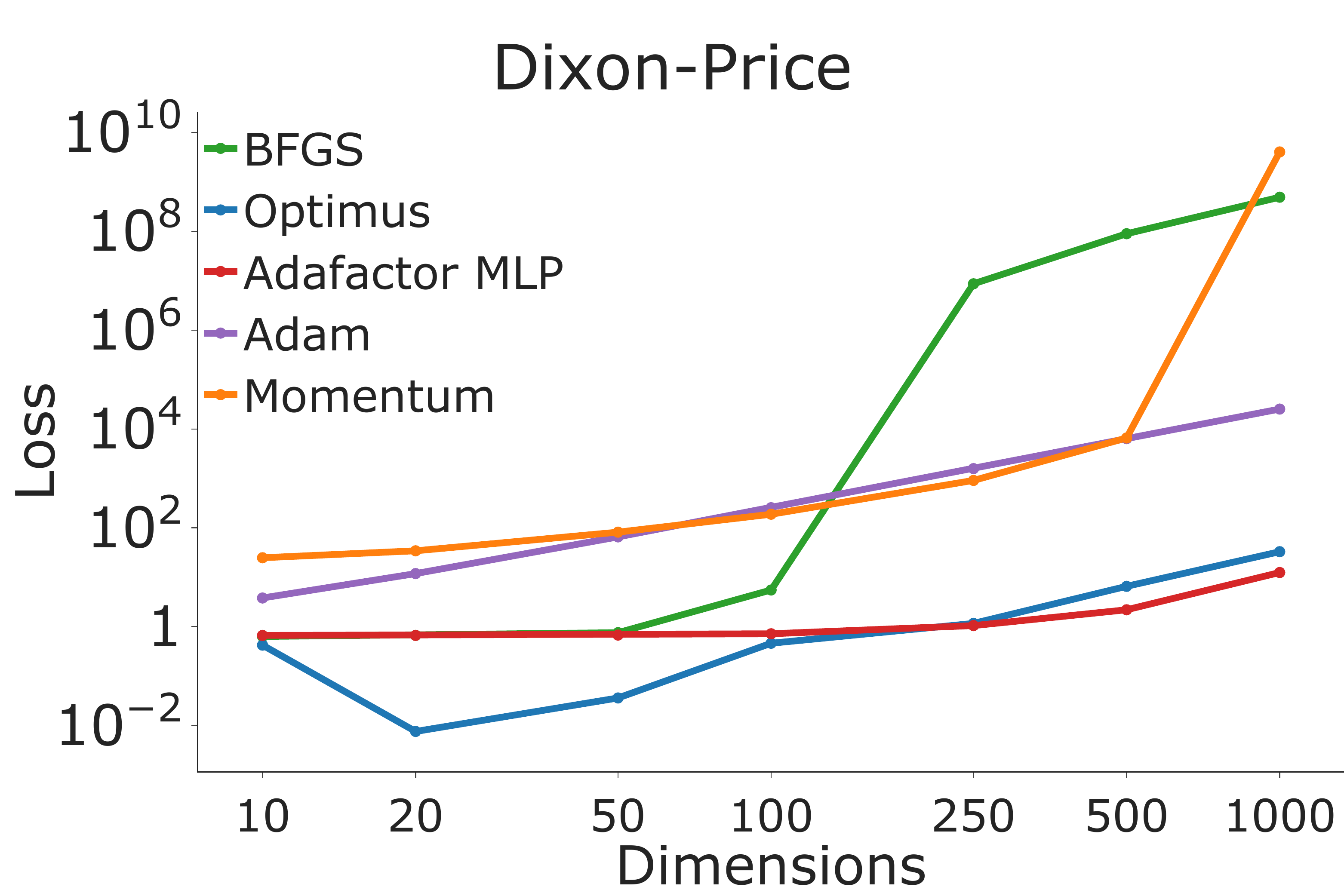}} &
\subfloat[Griwank]{\includegraphics[width = 0.30\linewidth]{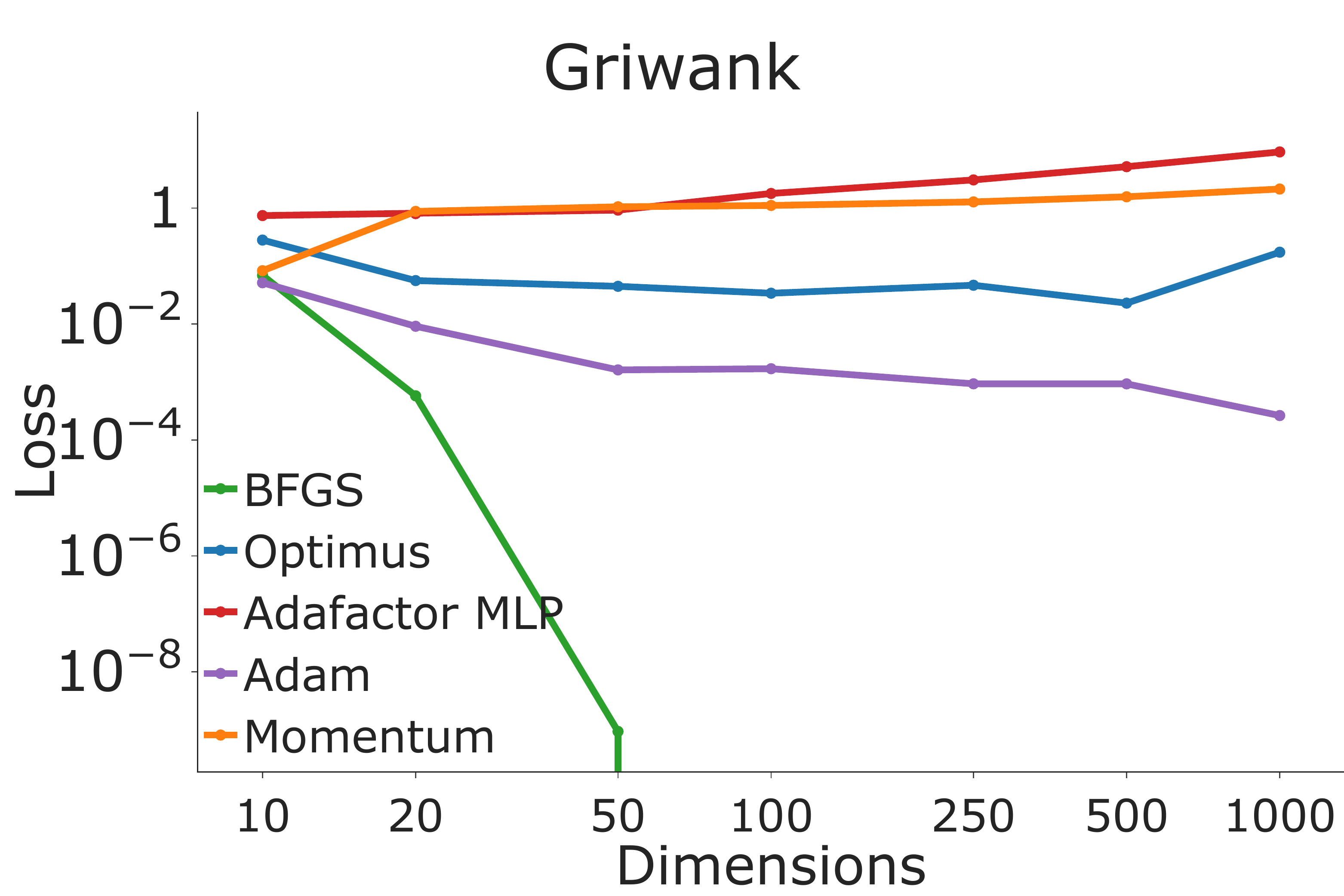}} \\
\subfloat[Levy]{\includegraphics[width = 0.30\linewidth]{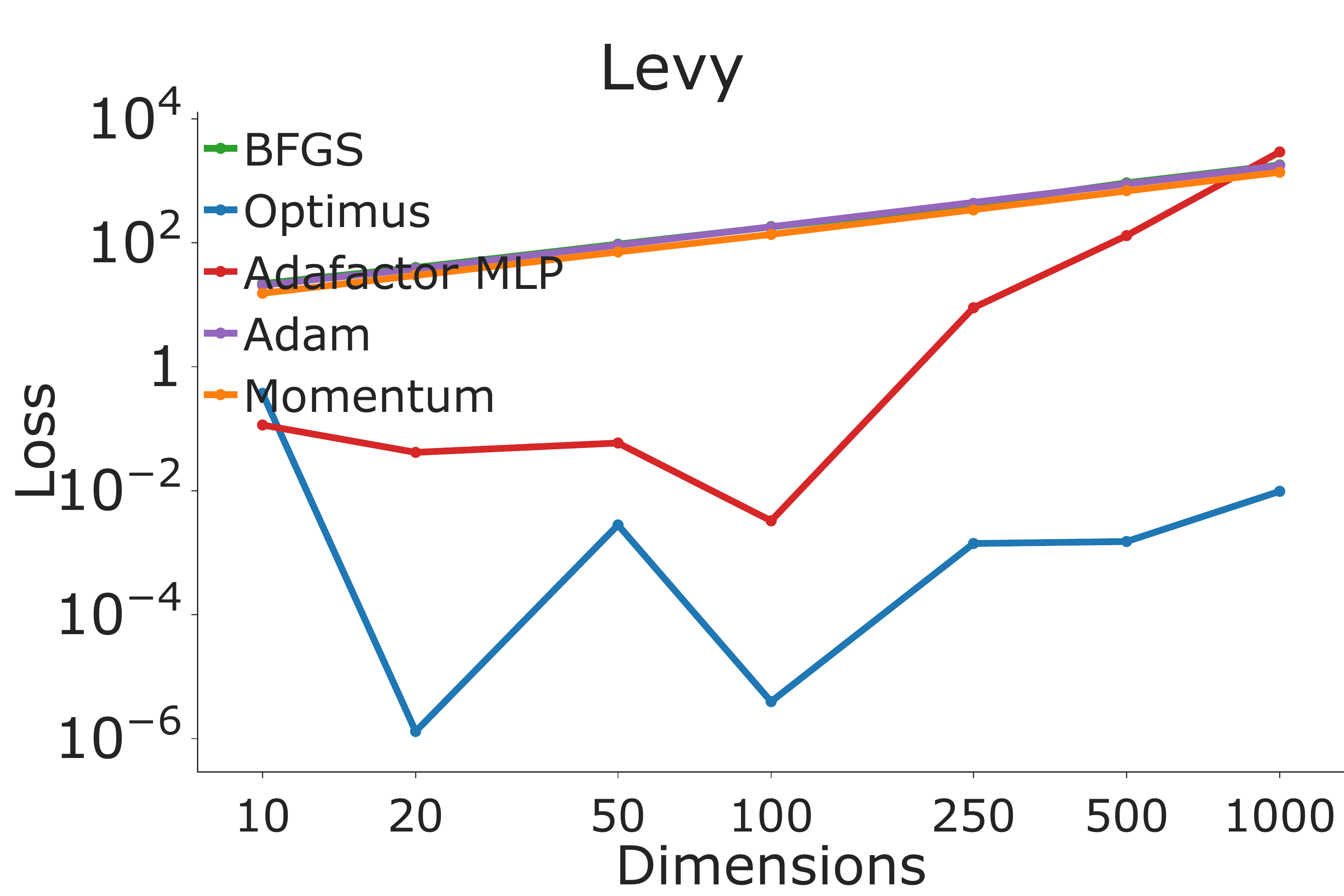}} &
\subfloat[Perm Function 0, d, beta]{\includegraphics[width = 0.30\linewidth]{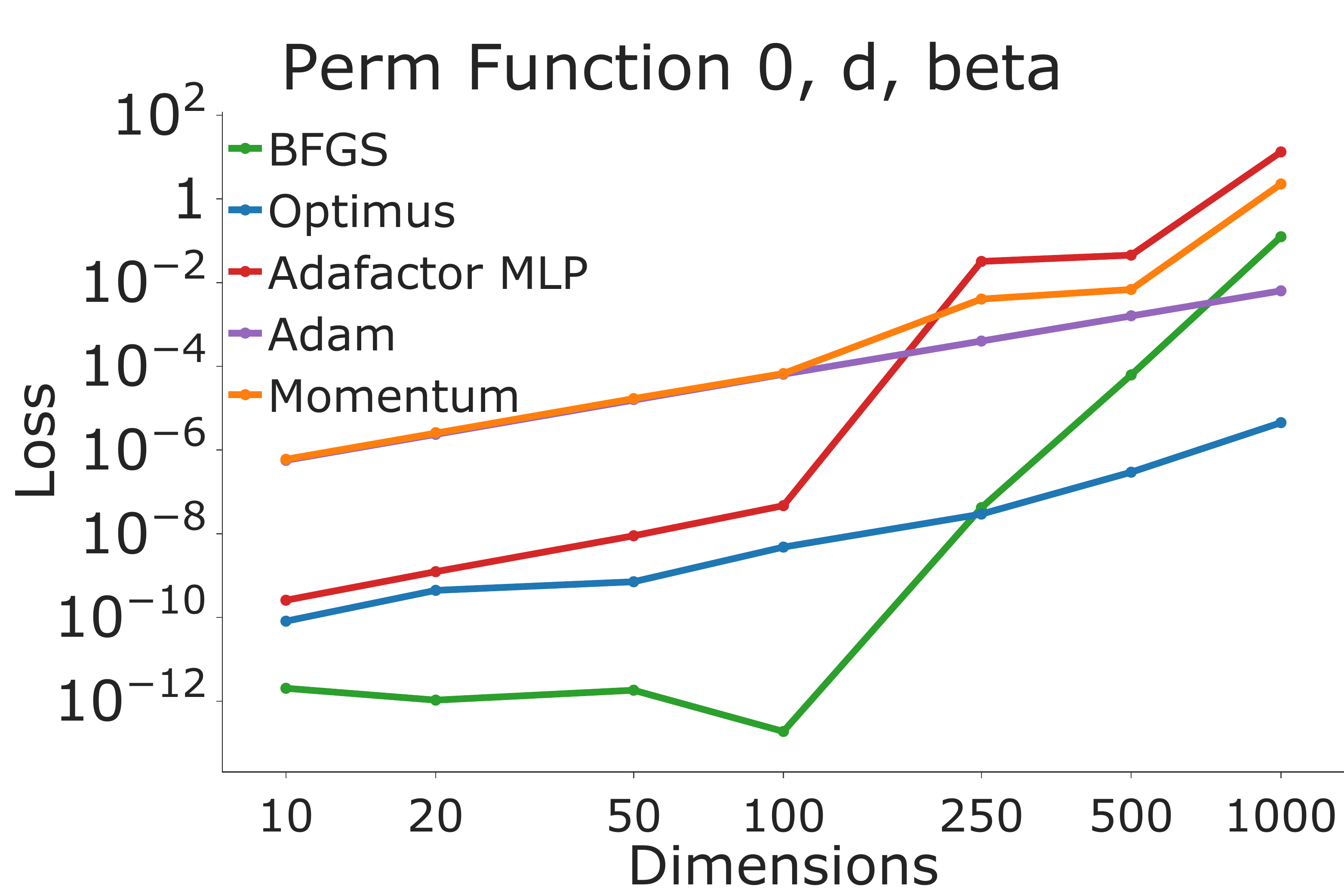}} &
\subfloat[Powel]{\includegraphics[width = 0.30\linewidth]{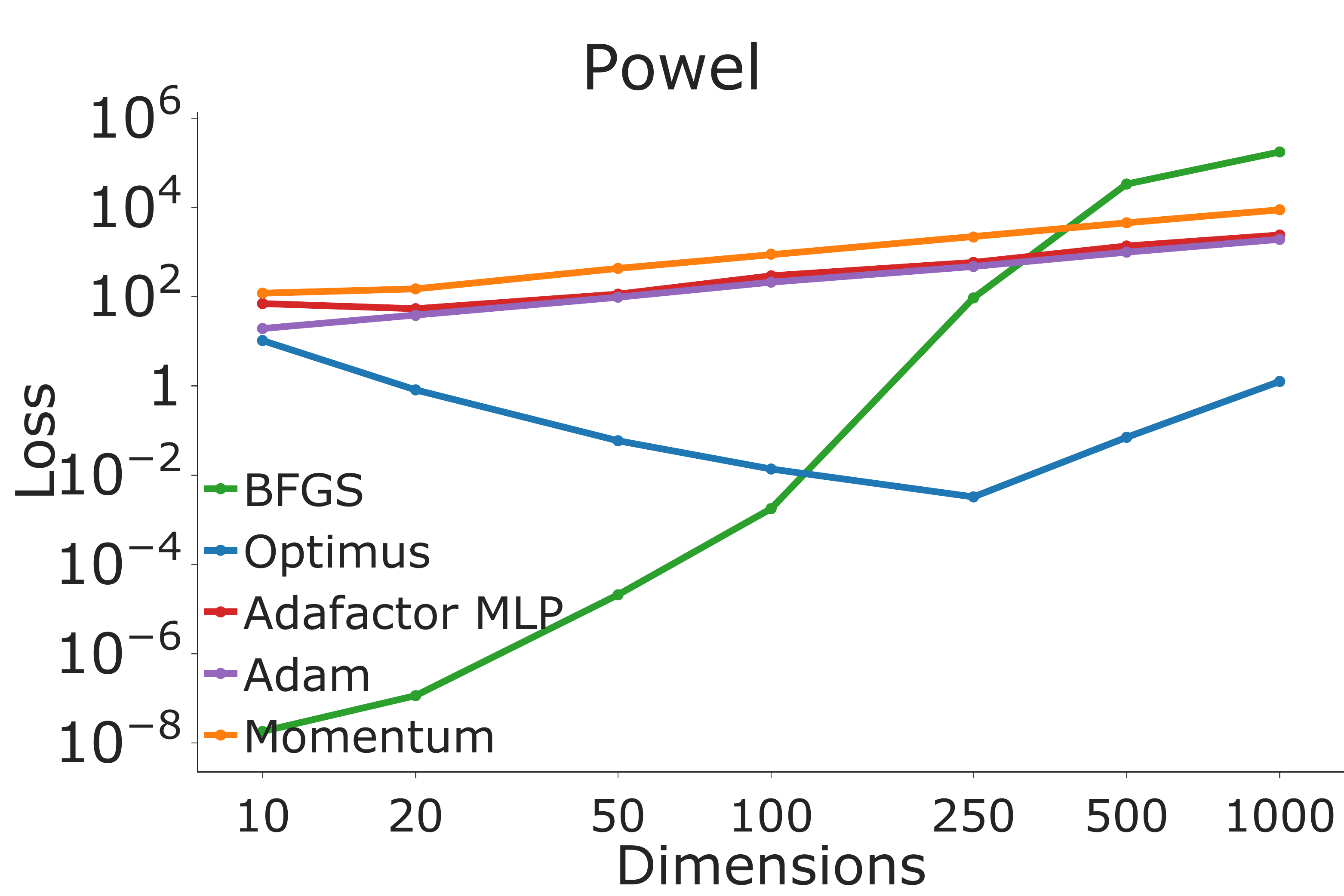}} \\
\subfloat[Rastrigin]{\includegraphics[width = 0.30\linewidth]{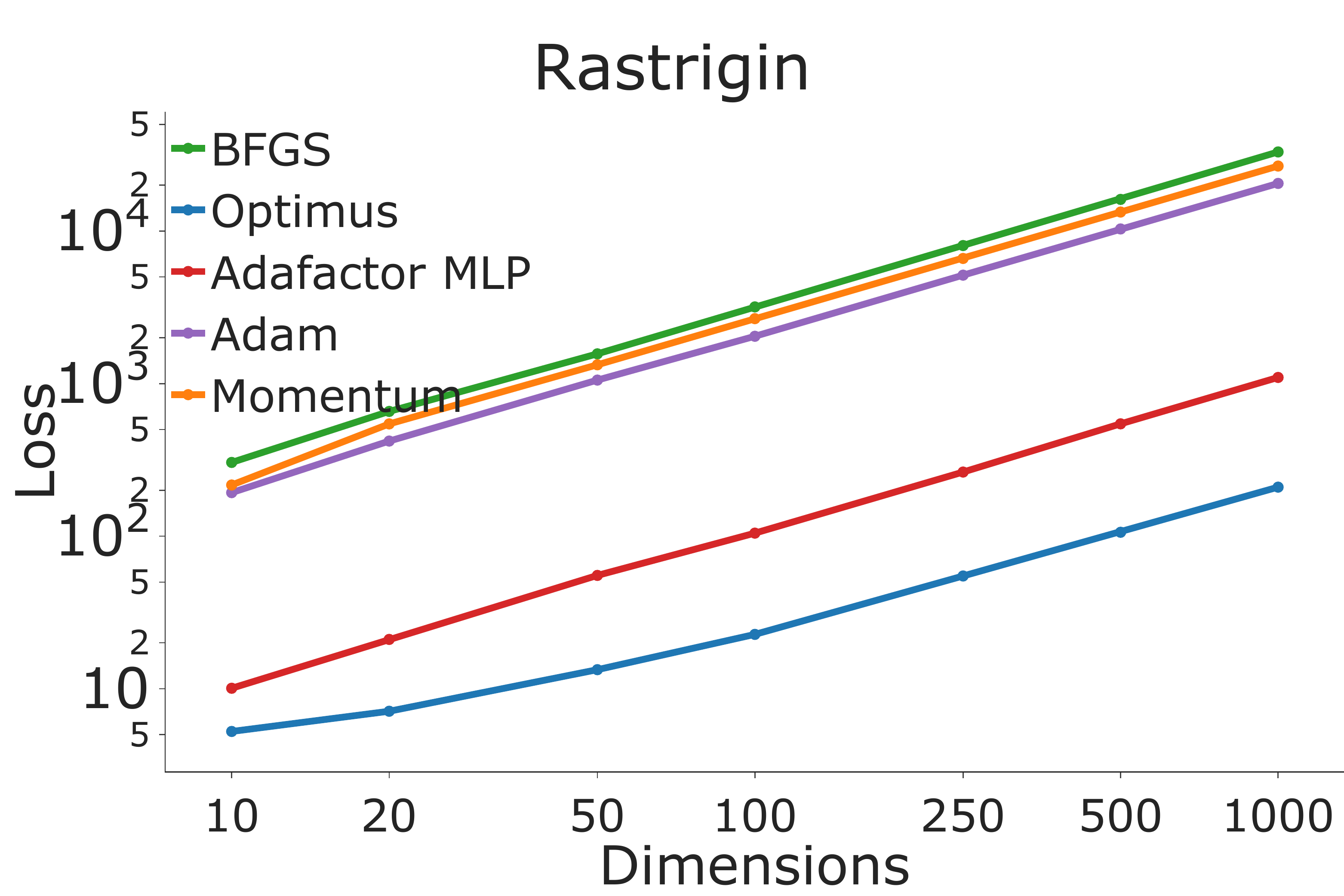}} &
\subfloat[Rosenbrock]{\includegraphics[width = 0.30\linewidth]{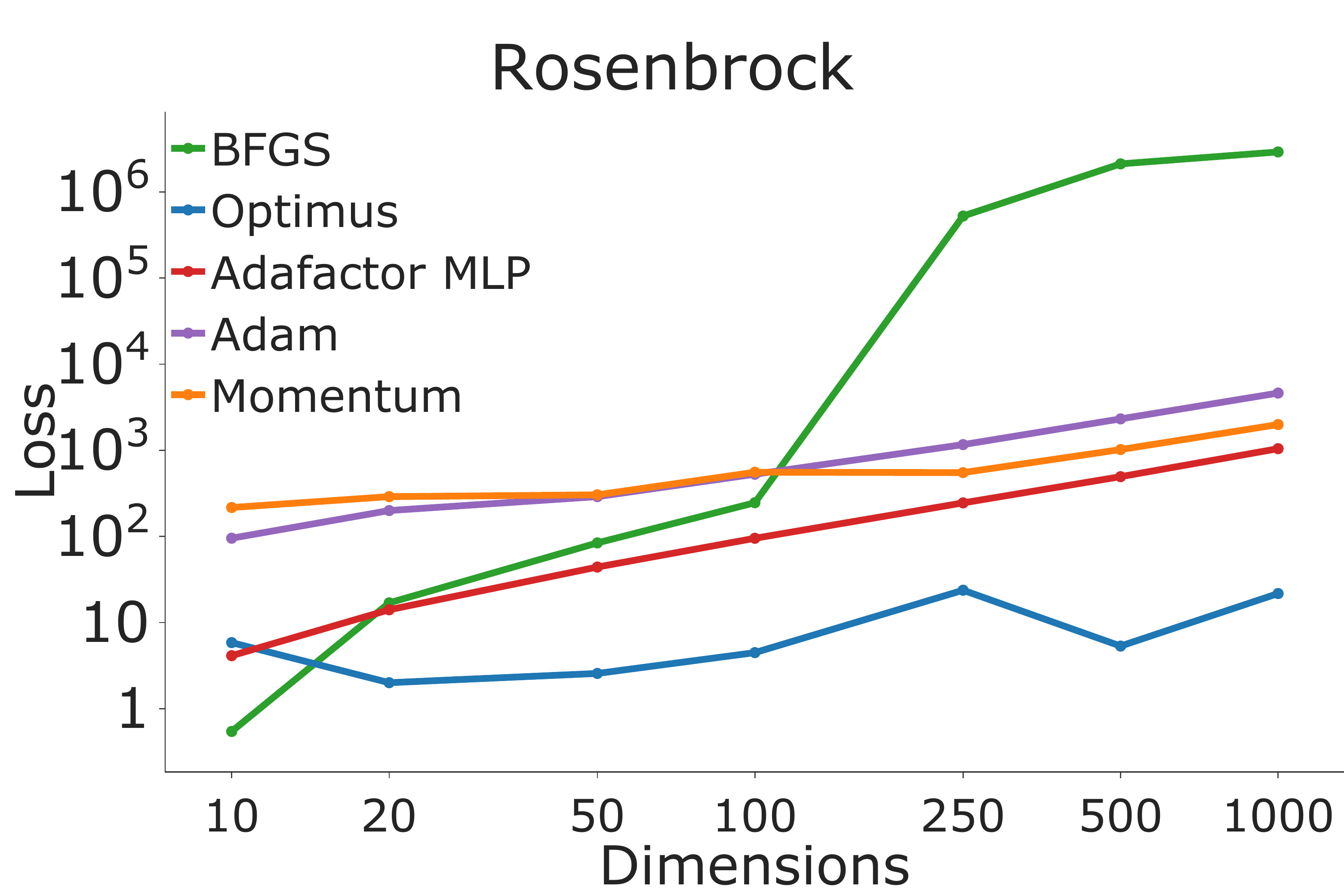}} &
\subfloat[Rotated Hyper-Ellipsoid]{\includegraphics[width = 0.30\linewidth]{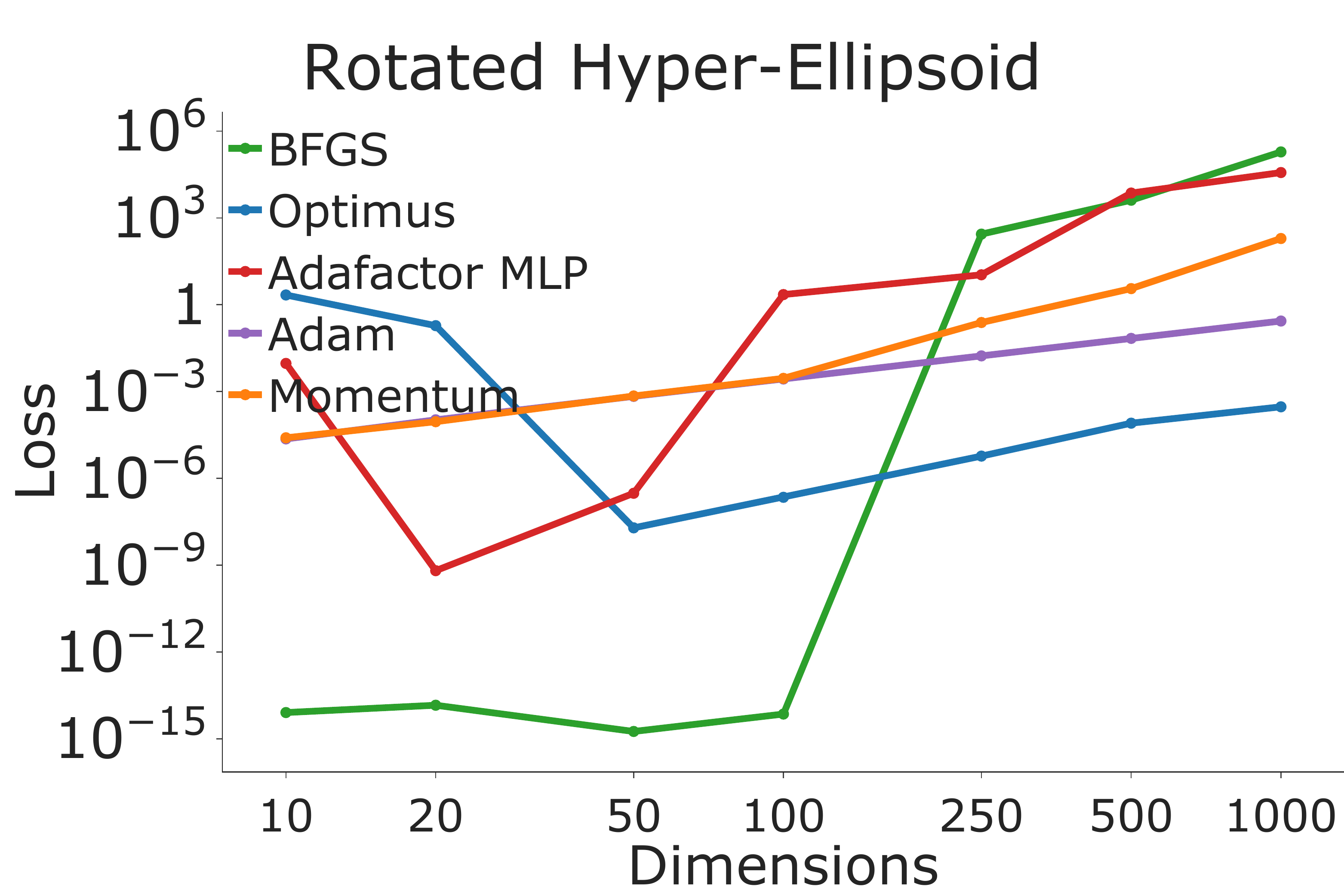}} \\
\subfloat[Sphere]{\includegraphics[width = 0.30\linewidth]{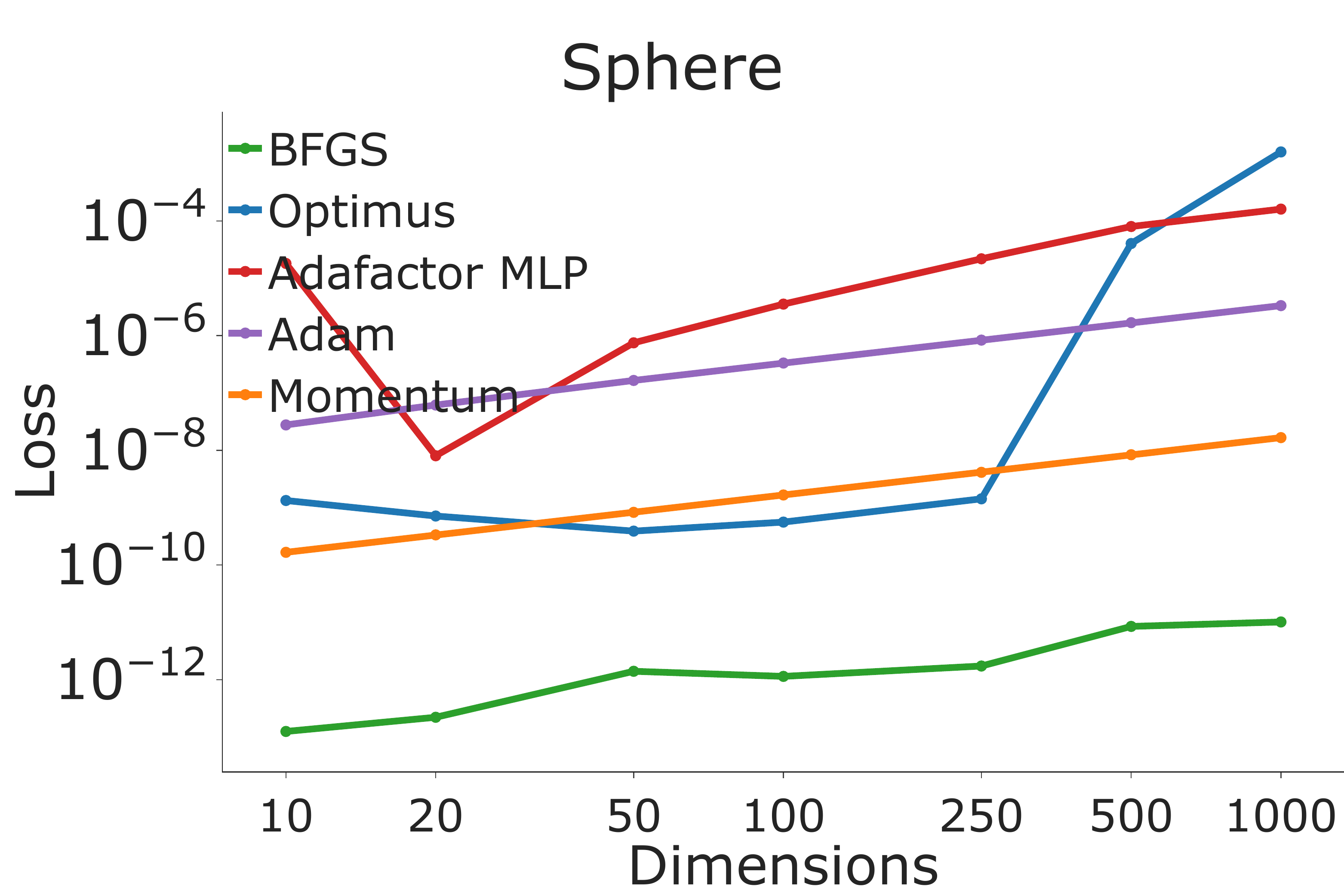}} &
\subfloat[Styblinski-Tang]{\includegraphics[width = 0.30\linewidth]{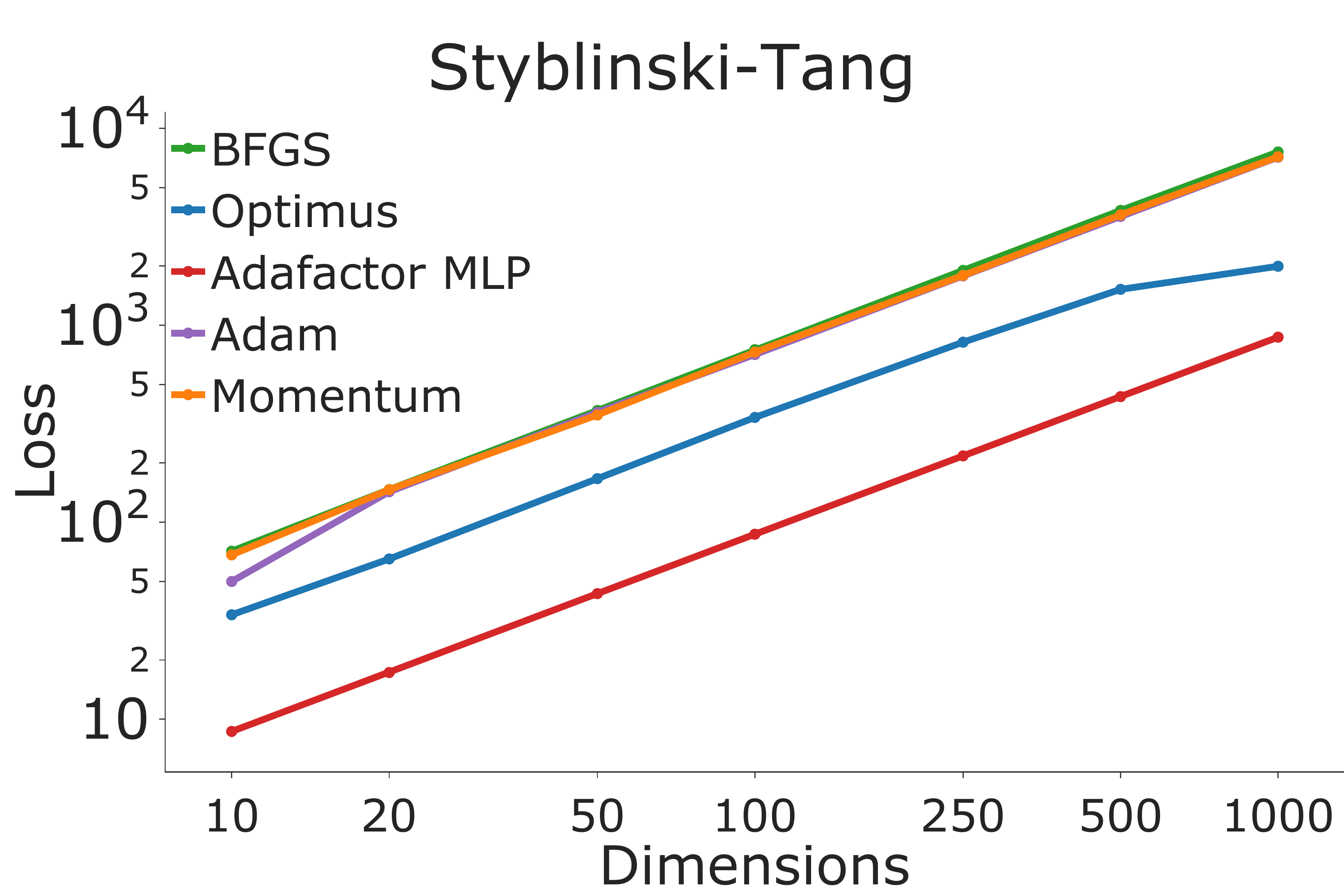}} &
\subfloat[Sum of Powers]{\includegraphics[width = 0.30\linewidth]{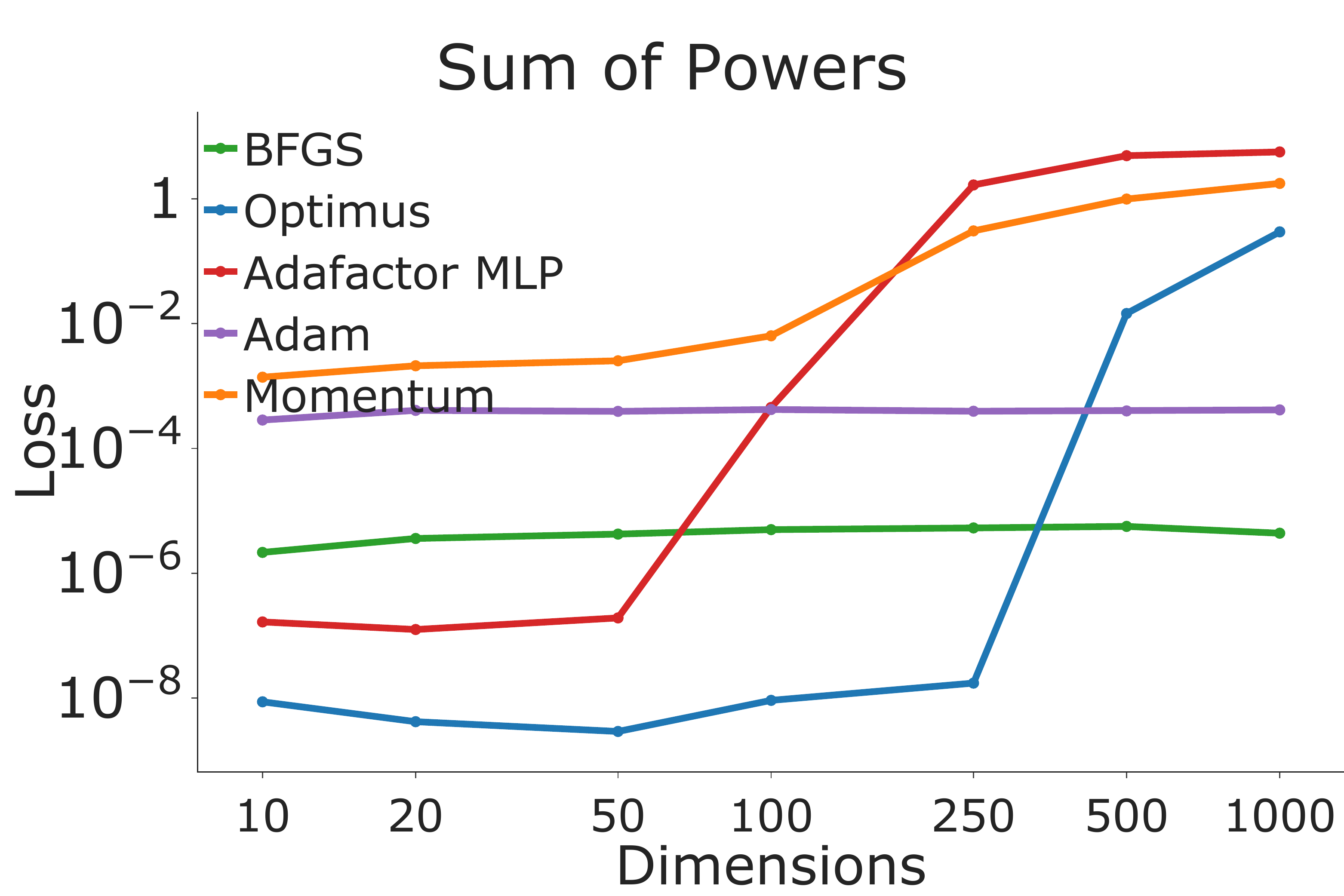}} \\
\subfloat[Sum of Squares]{\includegraphics[width = 0.30\linewidth]{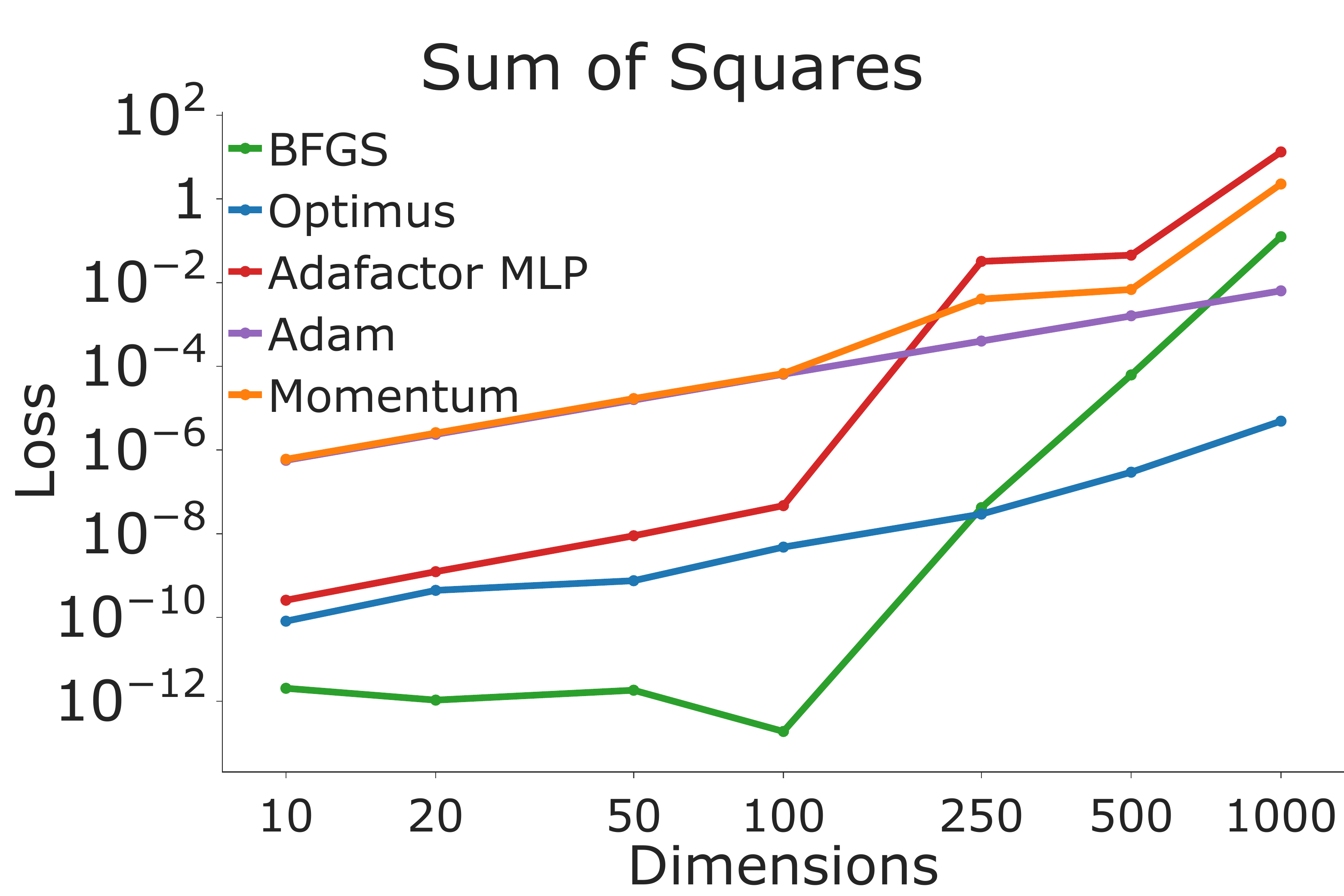}} &
\subfloat[Trid]{\includegraphics[width = 0.30\linewidth]{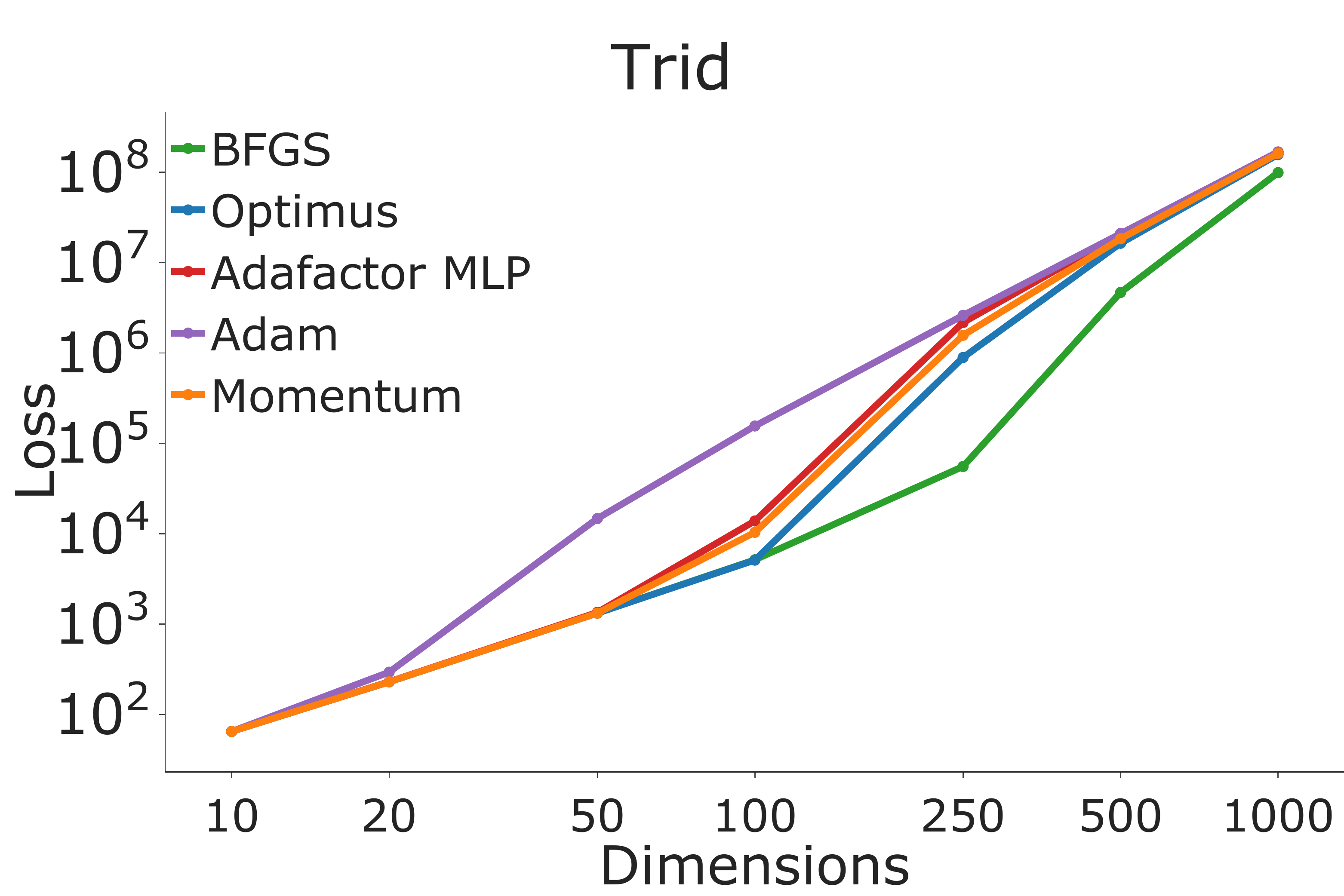}} &
\subfloat[Zakharov]{\includegraphics[width = 0.30\linewidth]{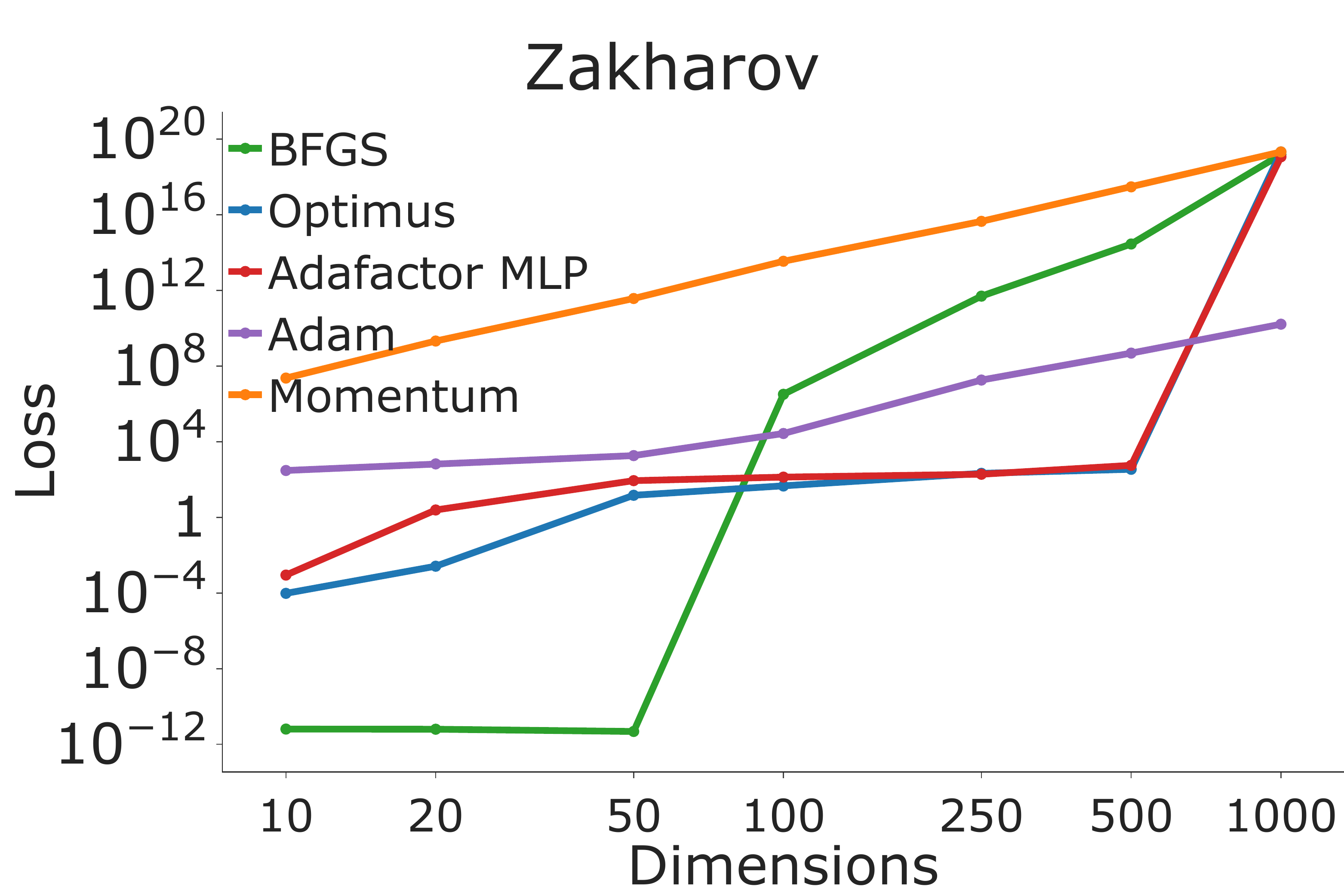}}

\end{tabular}
\caption{Evaluation results used to generate performance profile plot in Sec.~5.1 of the paper. Here we show results separately for each objective function. We plot objective value averaged over $64$ randomly initialized optimization runs on the y-axis and dimensionality of the objective function on the x-axis. Each method has a fixed budget of $200$ iterations.}
\label{fig:classical_function_results}
\end{figure*}
\begin{figure*}[p]
\centering
\begin{tabular}{ccc}
\subfloat[10d Rosenbrock]{\includegraphics[width = 0.31\linewidth]{{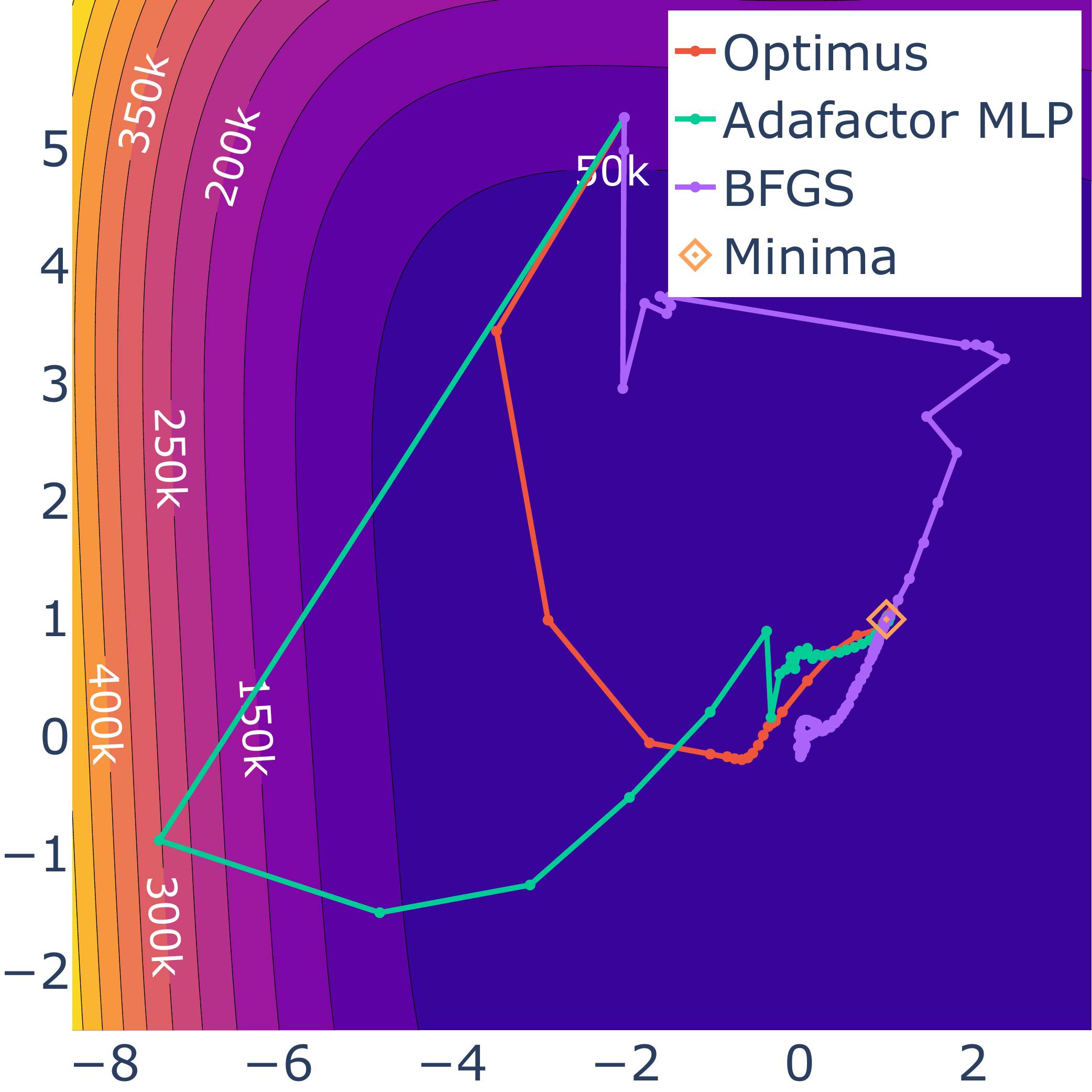}}} &
\subfloat[10d Rosenbrock]{\includegraphics[width = 0.31\linewidth]{{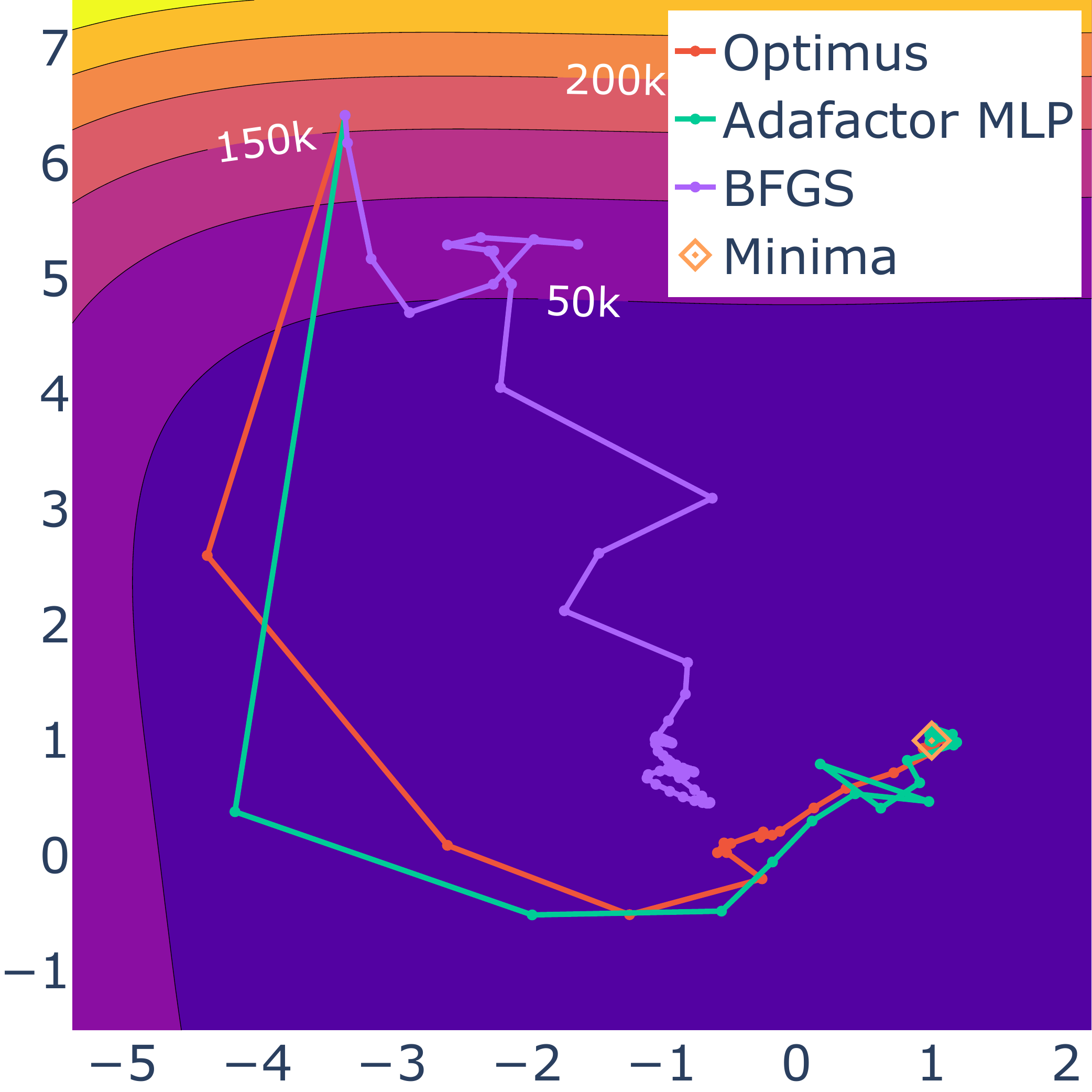}}} &
\subfloat[10d Rosenbrock]{\includegraphics[width = 0.31\linewidth]{{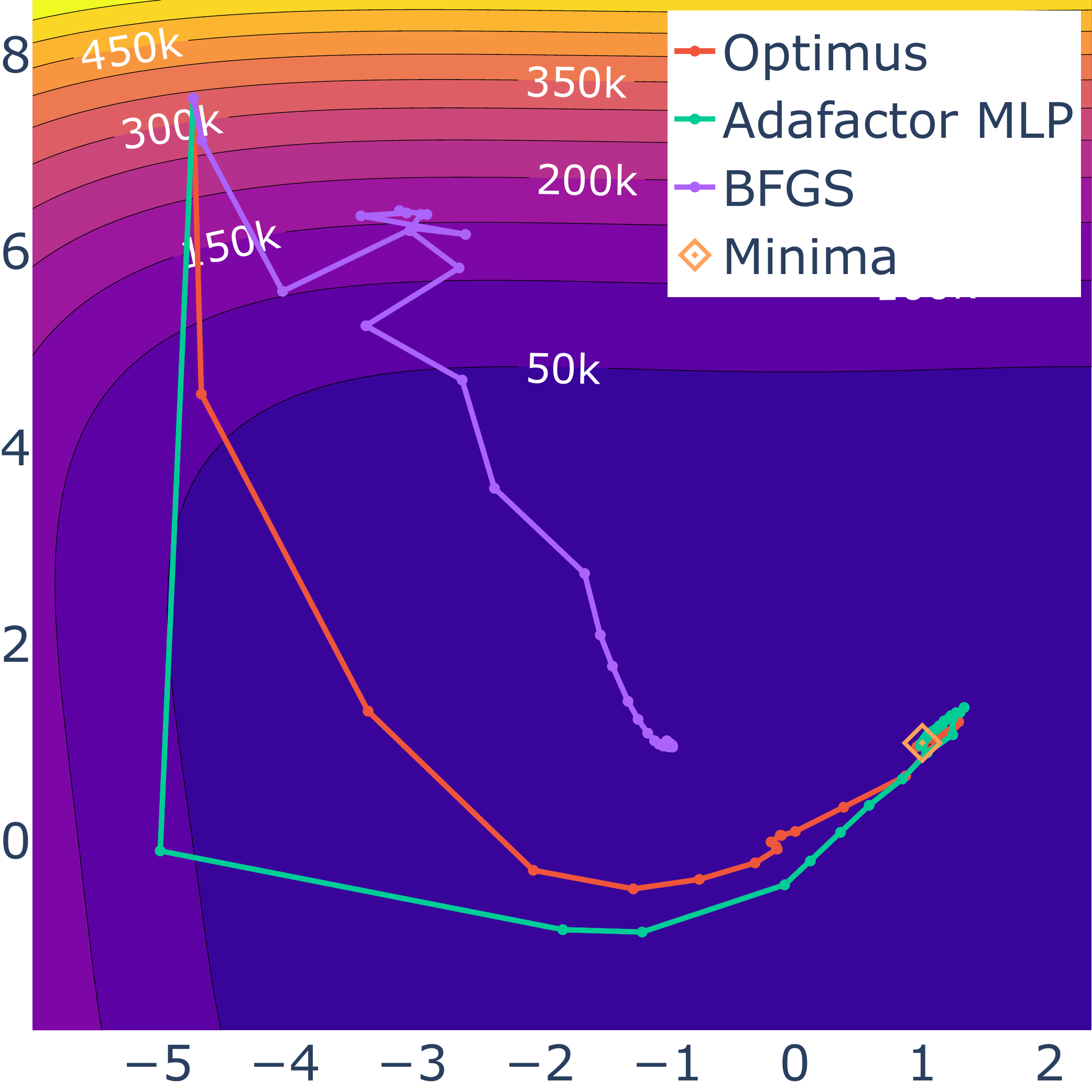}}} \\
\subfloat[100d Rosenbrock]{\includegraphics[width = 0.31\linewidth]{{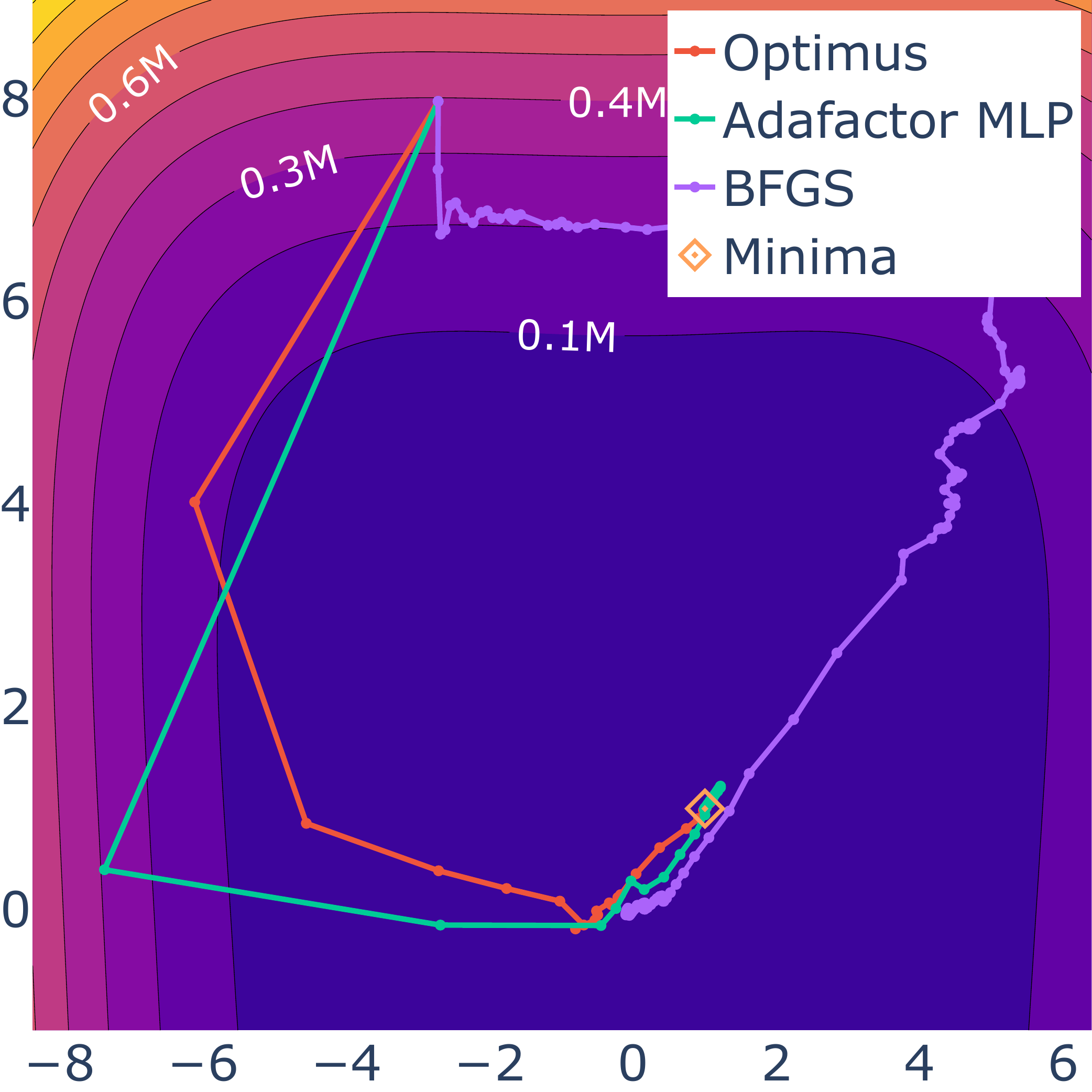}}} &
\subfloat[100d Rosenbrock]{\includegraphics[width = 0.31\linewidth]{{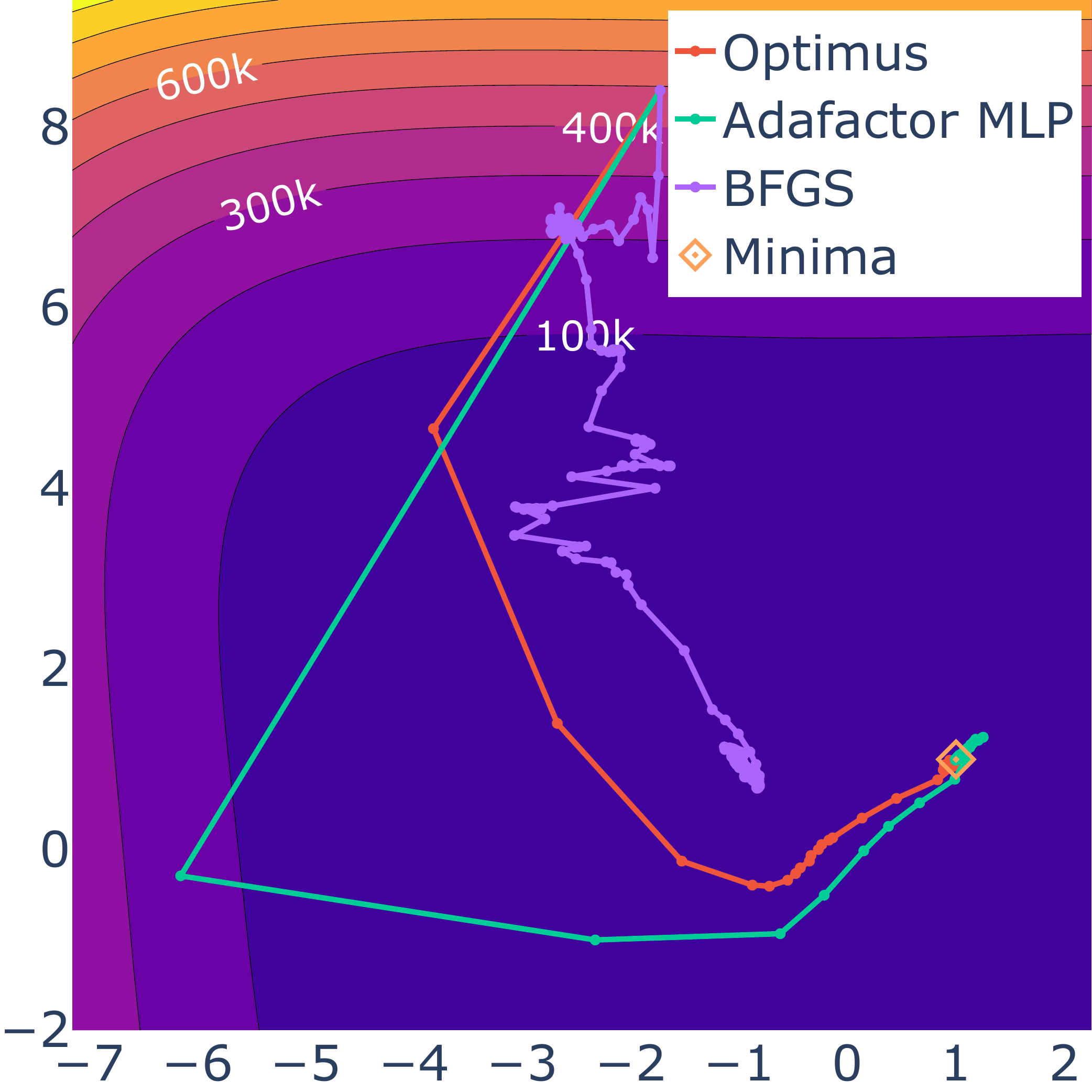}}} &
\subfloat[100d Rosenbrock]{\includegraphics[width = 0.31\linewidth]{{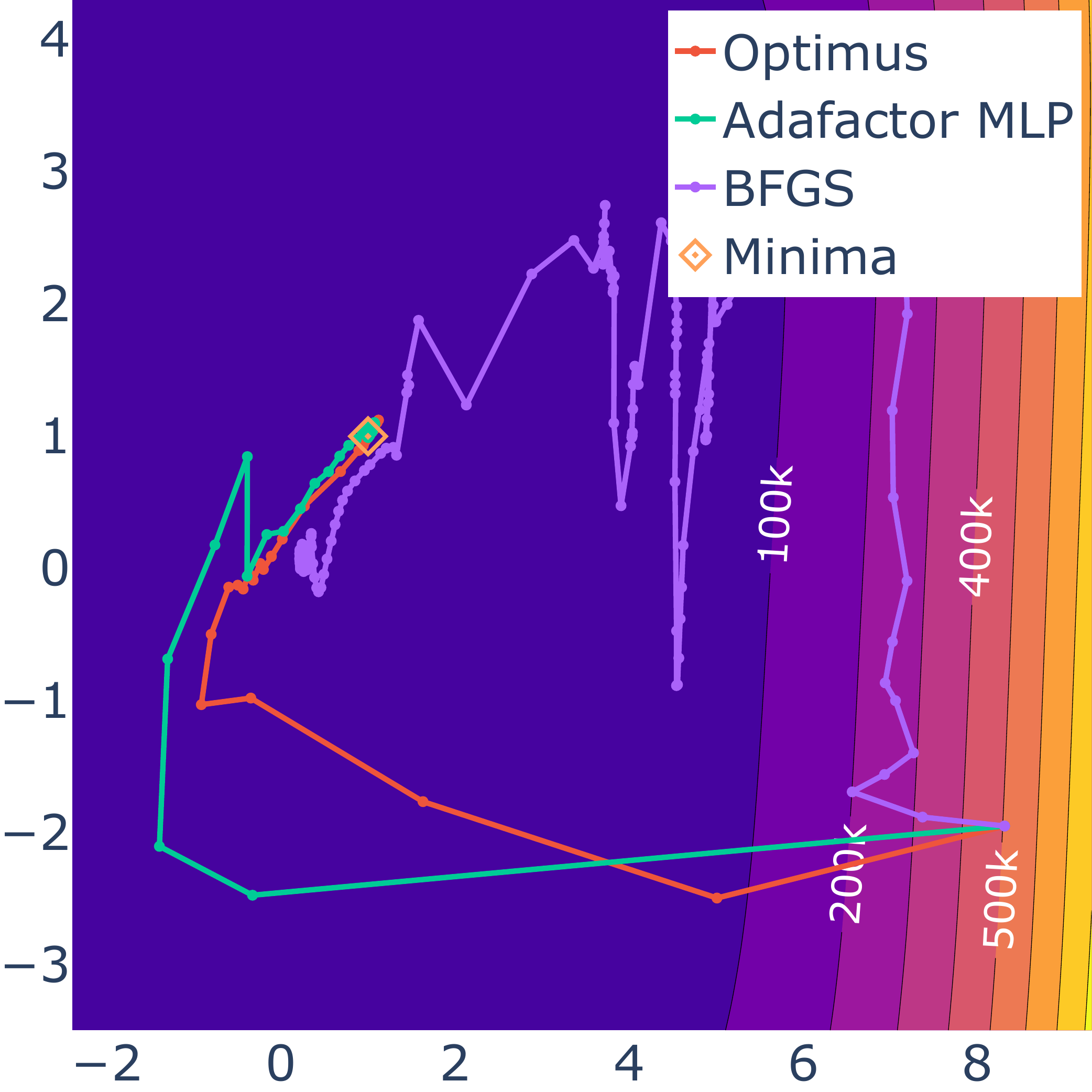}}} \\

\subfloat[1000d Rosenbrock]{\includegraphics[width = 0.31\linewidth]{{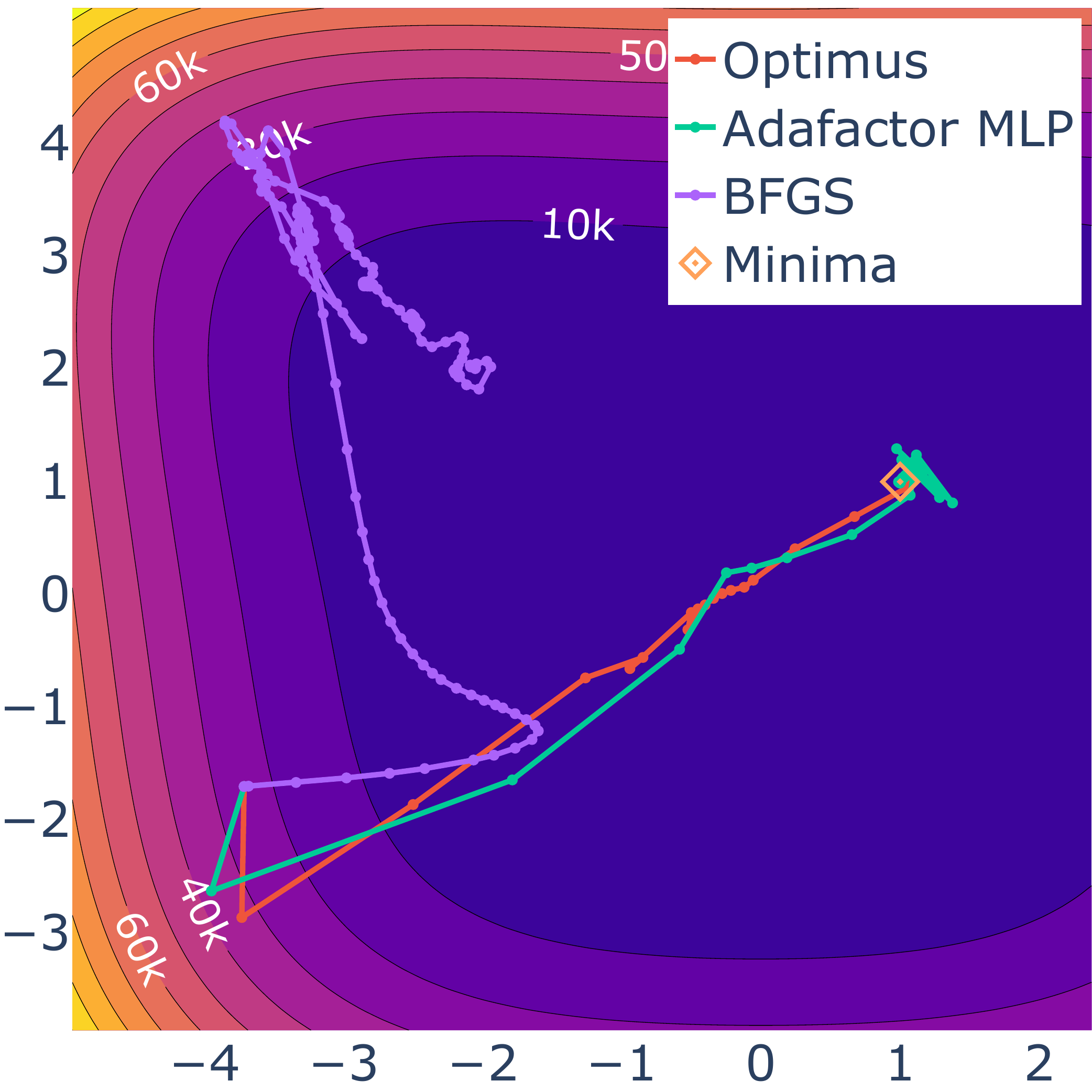}}} &
\subfloat[1000d Rosenbrock]{\includegraphics[width = 0.31\linewidth]{{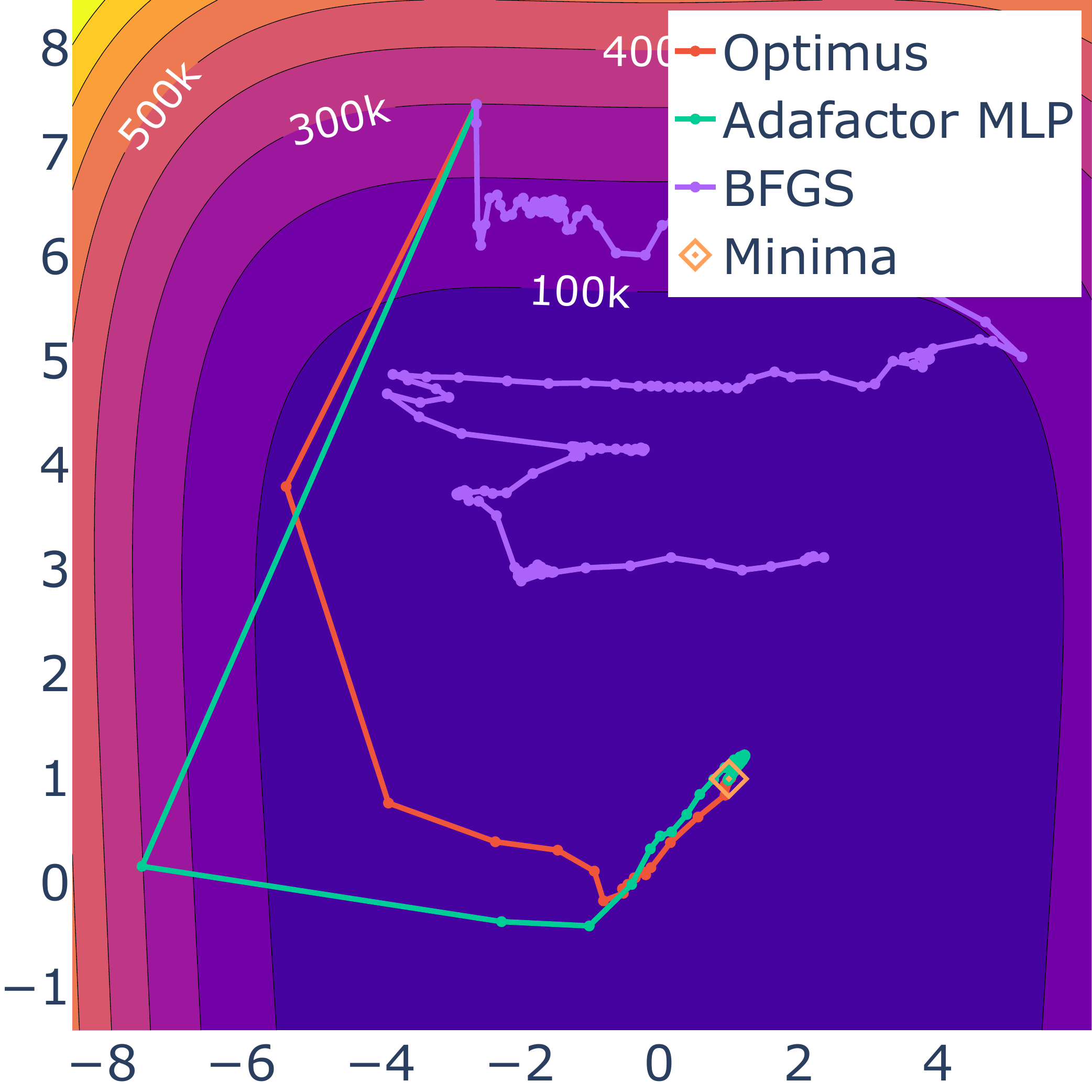}}} &
\subfloat[1000d Rosenbrock]{\includegraphics[width = 0.31\linewidth]{{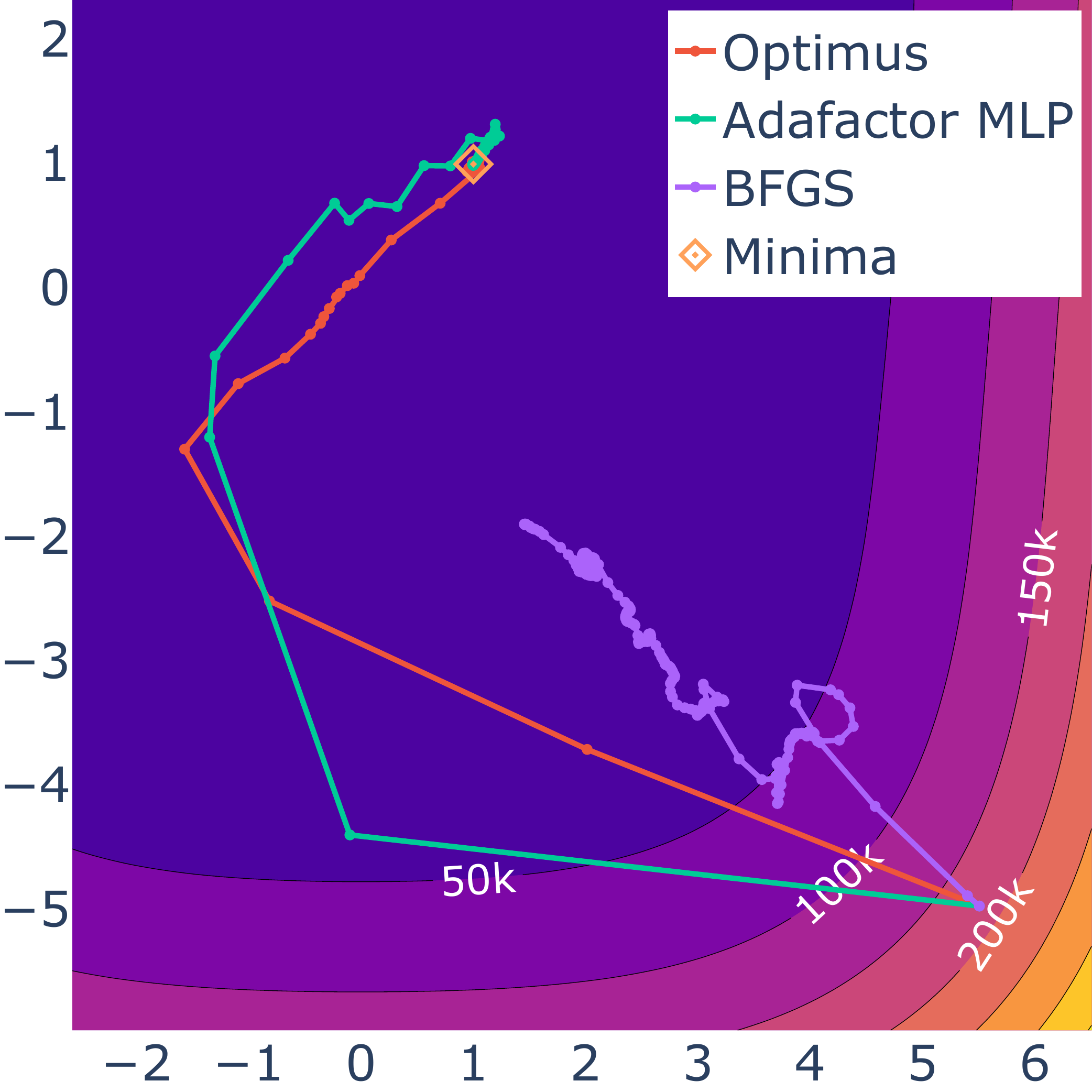}}}

\end{tabular}
\caption{Example trajectories on $N$-dimensional Rosenbrock functions. We visualize the trajectory for the first two dimensions.}
\label{fig:more_rosenbrock_trajectories}
\end{figure*}
\onecolumn

\begin{longtable}{|l|c|c|c|c|c|}
\caption{Full results for experiments on classical optimization functions. Each value is averaged over $64$ random initializations. We implement the functions according to the formulas presented in \cite{simulationlib}.}
\label{tab:full_classical_results} \\

\hline \multicolumn{1}{|c|}{\textbf{Function}} & \multicolumn{1}{c|}{\textbf{BFGS}} & \multicolumn{1}{c|}{\textbf{Optimus}} & \multicolumn{1}{c|}{\textbf{Adafactor MLP}}& \multicolumn{1}{c|}{\textbf{Adam}}&
\multicolumn{1}{c|}{\textbf{Momentum}}\\ \hline 
\endfirsthead

\multicolumn{6}{c}%
{{\bfseries \tablename\ \thetable{} -- continued from previous page}} \\
\hline \multicolumn{1}{|c|}{\textbf{Function}} & \multicolumn{1}{c|}{\textbf{BFGS}} & \multicolumn{1}{c|}{\textbf{Optimus}} & \multicolumn{1}{c|}{\textbf{Adafactor MLP}}& \multicolumn{1}{c|}{\textbf{Adam}}&
\multicolumn{1}{c|}{\textbf{Momentum}}\\ \hline 
\endhead

\hline \multicolumn{6}{|r|}{{Continued on next page}} \\ \hline
\endfoot

\hline \hline
\endlastfoot
Ackley 2d &  1.11e+01  &  6.65e+00  & \textbf{ 1.69e-01 } &  8.51e+00  &  8.99e+00  \\
Ackley 10d &  1.34e+01  &  7.16e-01  & \textbf{ 1.11e-01 } &  1.06e+01  &  1.31e+01  \\
Ackley 20d &  1.36e+01  & \textbf{ 2.12e-01 } &  2.68e-01  &  1.16e+01  &  1.34e+01  \\
Ackley 50d &  1.36e+01  & \textbf{ 5.46e-02 } &  5.19e-01  &  1.15e+01  &  1.34e+01  \\
Ackley 100d &  1.37e+01  & \textbf{ 6.80e-06 } &  9.02e-02  &  1.18e+01  &  1.36e+01  \\
Ackley 250d &  1.36e+01  &  3.65e-01  & \textbf{ 1.26e-01 } &  1.18e+01  &  1.36e+01  \\
Ackley 500d &  1.37e+01  &  1.38e+00  & \textbf{ 1.13e-01 } &  1.17e+01  &  1.37e+01  \\
Ackley 1000d &  1.37e+01  & \textbf{ 5.60e-05 } &  7.71e-02  &  1.16e+01  &  1.37e+01  \\
\hline
Dixon-Price 2d & \textbf{ 4.64e-13 } &  4.67e+00  &  8.86e-06  &  2.83e-01  &  5.42e+00  \\
Dixon-Price 10d &  6.35e-01  & \textbf{ 4.24e-01 } &  6.67e-01  &  3.81e+00  &  2.49e+01  \\
Dixon-Price 20d &  6.67e-01  & \textbf{ 7.60e-03 } &  6.68e-01  &  1.19e+01  &  3.44e+01  \\
Dixon-Price 50d &  7.58e-01  & \textbf{ 3.62e-02 } &  6.77e-01  &  6.56e+01  &  8.19e+01  \\
Dixon-Price 100d &  5.54e+00  & \textbf{ 4.63e-01 } &  7.20e-01  &  2.56e+02  &  1.89e+02  \\
Dixon-Price 250d &  8.67e+06  &  1.17e+00  & \textbf{ 1.05e+00 } &  1.58e+03  &  9.12e+02  \\
Dixon-Price 500d &  8.93e+07  &  6.57e+00  & \textbf{ 2.20e+00 } &  6.39e+03  &  6.60e+03  \\
Dixon-Price 1000d &  4.90e+08  &  3.30e+01  & \textbf{ 1.25e+01 } &  2.53e+04  &  4.04e+09  \\
\hline
Griwank 2d &  2.00e-02  &  9.04e-01  &  1.43e-01  &  1.96e-02  & \textbf{ 1.91e-02 } \\
Griwank 10d &  6.92e-02  &  2.81e-01  &  7.45e-01  & \textbf{ 5.16e-02 } &  8.33e-02  \\
Griwank 20d & \textbf{ 5.78e-04 } &  5.63e-02  &  8.05e-01  &  9.16e-03  &  8.78e-01  \\
Griwank 50d & \textbf{ 9.31e-10 } &  4.48e-02  &  9.25e-01  &  1.62e-03  &  1.06e+00  \\
Griwank 100d & \textbf{ 0.00e+00 } &  3.41e-02  &  1.79e+00  &  1.69e-03  &  1.11e+00  \\
Griwank 250d & \textbf{ 0.00e+00 } &  4.68e-02  &  3.07e+00  &  9.31e-04  &  1.28e+00  \\
Griwank 500d & \textbf{ 0.00e+00 } &  2.30e-02  &  5.20e+00  &  9.29e-04  &  1.57e+00  \\
Griwank 1000d & \textbf{ 0.00e+00 } &  1.74e-01  &  9.34e+00  &  2.64e-04  &  2.14e+00  \\
\hline
Levy 2d &  8.36e+00  &  5.65e+00  & \textbf{ 3.74e+00 } &  8.40e+00  &  3.80e+00  \\
Levy 10d &  2.20e+01  &  3.72e-01  & \textbf{ 1.15e-01 } &  2.06e+01  &  1.53e+01  \\
Levy 20d &  4.00e+01  & \textbf{ 1.31e-06 } &  4.16e-02  &  3.81e+01  &  3.03e+01  \\
Levy 50d &  9.51e+01  & \textbf{ 2.81e-03 } &  5.89e-02  &  9.17e+01  &  7.12e+01  \\
Levy 100d &  1.83e+02  & \textbf{ 3.96e-06 } &  3.27e-03  &  1.78e+02  &  1.37e+02  \\
Levy 250d &  4.20e+02  & \textbf{ 1.41e-03 } &  8.96e+00  &  4.34e+02  &  3.39e+02  \\
Levy 500d &  9.28e+02  & \textbf{ 1.52e-03 } &  1.31e+02  &  8.89e+02  &  6.95e+02  \\
Levy 1000d &  1.81e+03  & \textbf{ 9.78e-03 } &  2.93e+03  &  1.76e+03  &  1.38e+03  \\
\hline
Perm Function 0, d, beta 2d & \textbf{ 1.76e-12 } &  1.16e-01  &  1.07e-01  &  9.96e-09  &  5.65e-09  \\
Perm Function 0, d, beta 10d & \textbf{ 2.03e-12 } &  8.17e-11  &  2.58e-10  &  5.69e-07  &  6.02e-07  \\
Perm Function 0, d, beta 20d & \textbf{ 1.06e-12 } &  4.45e-10  &  1.24e-09  &  2.37e-06  &  2.60e-06  \\
Perm Function 0, d, beta 50d & \textbf{ 1.83e-12 } &  7.14e-10  &  8.96e-09  &  1.61e-05  &  1.70e-05  \\
Perm Function 0, d, beta 100d & \textbf{ 1.89e-13 } &  4.80e-09  &  4.70e-08  &  6.49e-05  &  6.73e-05  \\
Perm Function 0, d, beta 250d &  4.21e-08  & \textbf{ 2.95e-08 } &  3.21e-02  &  4.02e-04  &  4.05e-03  \\
Perm Function 0, d, beta 500d &  6.24e-05  & \textbf{ 2.96e-07 } &  4.53e-02  &  1.61e-03  &  6.86e-03  \\
Perm Function 0, d, beta 1000d &  1.25e-01  & \textbf{ 4.52e-06 } &  1.33e+01  &  6.36e-03  &  2.27e+00  \\
\hline
Powel 2d & \textbf{ 1.08e-08 } &  4.29e-01  &  5.16e+00  &  7.83e+00  &  3.43e+01  \\
Powel 10d & \textbf{ 1.81e-08 } &  1.04e+01  &  7.02e+01  &  1.94e+01  &  1.20e+02  \\
Powel 20d & \textbf{ 1.15e-07 } &  8.17e-01  &  5.39e+01  &  3.82e+01  &  1.50e+02  \\
Powel 50d & \textbf{ 2.09e-05 } &  5.92e-02  &  1.14e+02  &  9.74e+01  &  4.31e+02  \\
Powel 100d & \textbf{ 1.78e-03 } &  1.38e-02  &  2.94e+02  &  2.12e+02  &  8.98e+02  \\
Powel 250d &  9.41e+01  & \textbf{ 3.26e-03 } &  5.87e+02  &  4.77e+02  &  2.18e+03  \\
Powel 500d &  3.35e+04  & \textbf{ 7.06e-02 } &  1.37e+03  &  9.98e+02  &  4.54e+03  \\
Powel 1000d &  1.76e+05  & \textbf{ 1.26e+00 } &  2.41e+03  &  1.93e+03  &  8.88e+03  \\
\hline
Rastrigin 2d &  6.17e+01  &  3.30e+00  & \textbf{ 2.85e+00 } &  3.99e+01  &  5.59e+01  \\
Rastrigin 10d &  3.04e+02  & \textbf{ 5.24e+00 } &  1.01e+01  &  1.93e+02  &  2.16e+02  \\
Rastrigin 20d &  6.57e+02  & \textbf{ 7.12e+00 } &  2.10e+01  &  4.20e+02  &  5.44e+02  \\
Rastrigin 50d &  1.57e+03  & \textbf{ 1.33e+01 } &  5.53e+01  &  1.05e+03  &  1.33e+03  \\
Rastrigin 100d &  3.18e+03  & \textbf{ 2.27e+01 } &  1.05e+02  &  2.04e+03  &  2.66e+03  \\
Rastrigin 250d &  8.03e+03  & \textbf{ 5.48e+01 } &  2.63e+02  &  5.14e+03  &  6.61e+03  \\
Rastrigin 500d &  1.61e+04  & \textbf{ 1.06e+02 } &  5.45e+02  &  1.03e+04  &  1.33e+04  \\
Rastrigin 1000d &  3.30e+04  & \textbf{ 2.10e+02 } &  1.10e+03  &  2.05e+04  &  2.67e+04  \\
\hline
Rosenbrock 2d & \textbf{ 5.07e-13 } &  3.48e+01  &  5.06e-01  &  3.58e+00  &  5.17e+00  \\
Rosenbrock 10d & \textbf{ 5.43e-01 } &  5.84e+00  &  4.12e+00  &  9.52e+01  &  2.16e+02  \\
Rosenbrock 20d &  1.69e+01  & \textbf{ 2.00e+00 } &  1.41e+01  &  1.99e+02  &  2.89e+02  \\
Rosenbrock 50d &  8.40e+01  & \textbf{ 2.56e+00 } &  4.40e+01  &  2.89e+02  &  3.04e+02  \\
Rosenbrock 100d &  2.46e+02  & \textbf{ 4.47e+00 } &  9.50e+01  &  5.26e+02  &  5.56e+02  \\
Rosenbrock 250d &  5.24e+05  & \textbf{ 2.37e+01 } &  2.45e+02  &  1.16e+03  &  5.50e+02  \\
Rosenbrock 500d &  2.11e+06  & \textbf{ 5.33e+00 } &  4.93e+02  &  2.31e+03  &  1.02e+03  \\
Rosenbrock 1000d &  2.91e+06  & \textbf{ 2.17e+01 } &  1.04e+03  &  4.61e+03  &  1.99e+03  \\
\hline
Rotated Hyper-Ellipsoid 2d & \textbf{ 2.36e-14 } &  2.59e+01  &  1.74e+00  &  3.81e-07  &  5.32e-08  \\
Rotated Hyper-Ellipsoid 10d & \textbf{ 8.08e-15 } &  2.18e+00  &  9.38e-03  &  2.31e-05  &  2.53e-05  \\
Rotated Hyper-Ellipsoid 20d & \textbf{ 1.45e-14 } &  1.89e-01  &  6.46e-10  &  1.05e-04  &  9.07e-05  \\
Rotated Hyper-Ellipsoid 50d & \textbf{ 1.80e-15 } &  1.95e-08  &  3.06e-07  &  6.81e-04  &  7.03e-04  \\
Rotated Hyper-Ellipsoid 100d & \textbf{ 7.18e-15 } &  2.22e-07  &  2.24e+00  &  2.71e-03  &  2.86e-03  \\
Rotated Hyper-Ellipsoid 250d &  2.76e+02  & \textbf{ 5.88e-06 } &  1.08e+01  &  1.71e-02  &  2.43e-01  \\
Rotated Hyper-Ellipsoid 500d &  4.09e+03  & \textbf{ 8.07e-05 } &  7.24e+03  &  6.87e-02  &  3.61e+00  \\
Rotated Hyper-Ellipsoid 1000d &  1.91e+05  & \textbf{ 2.97e-04 } &  3.70e+04  &  2.73e-01  &  1.94e+02  \\
\hline
Sphere 2d & \textbf{ 7.07e-14 } &  4.24e-04  &  3.22e-03  &  1.84e-09  &  3.68e-11  \\
Sphere 10d & \textbf{ 1.25e-13 } &  1.33e-09  &  1.81e-05  &  2.77e-08  &  1.66e-10  \\
Sphere 20d & \textbf{ 2.21e-13 } &  7.14e-10  &  8.02e-09  &  6.15e-08  &  3.35e-10  \\
Sphere 50d & \textbf{ 1.40e-12 } &  3.89e-10  &  7.48e-07  &  1.65e-07  &  8.28e-10  \\
Sphere 100d & \textbf{ 1.14e-12 } &  5.59e-10  &  3.54e-06  &  3.32e-07  &  1.66e-09  \\
Sphere 250d & \textbf{ 1.73e-12 } &  1.42e-09  &  2.19e-05  &  8.34e-07  &  4.13e-09  \\
Sphere 500d & \textbf{ 8.48e-12 } &  4.04e-05  &  7.99e-05  &  1.68e-06  &  8.37e-09  \\
Sphere 1000d & \textbf{ 1.01e-11 } &  1.60e-03  &  1.61e-04  &  3.33e-06  &  1.66e-08  \\
\hline
Styblinski-Tang 2d &  1.40e+01  &  4.59e+00  & \textbf{ 3.40e+00 } &  1.10e+01  &  1.04e+01  \\
Styblinski-Tang 10d &  7.13e+01  &  3.39e+01  & \textbf{ 8.66e+00 } &  5.00e+01  &  6.83e+01  \\
Styblinski-Tang 20d &  1.47e+02  &  6.51e+01  & \textbf{ 1.73e+01 } &  1.43e+02  &  1.46e+02  \\
Styblinski-Tang 50d &  3.69e+02  &  1.66e+02  & \textbf{ 4.34e+01 } &  3.61e+02  &  3.50e+02  \\
Styblinski-Tang 100d &  7.52e+02  &  3.40e+02  & \textbf{ 8.69e+01 } &  7.10e+02  &  7.30e+02  \\
Styblinski-Tang 250d &  1.90e+03  &  8.20e+02  & \textbf{ 2.17e+02 } &  1.78e+03  &  1.79e+03  \\
Styblinski-Tang 500d &  3.83e+03  &  1.52e+03  & \textbf{ 4.34e+02 } &  3.57e+03  &  3.64e+03  \\
Styblinski-Tang 1000d &  7.59e+03  &  1.99e+03  & \textbf{ 8.70e+02 } &  7.13e+03  &  7.17e+03  \\
\hline
Sum of Powers 2d &  3.86e-08  &  2.16e-08  & \textbf{ 3.42e-09 } &  2.14e-06  &  3.96e-05  \\
Sum of Powers 10d &  2.16e-06  & \textbf{ 8.71e-09 } &  1.66e-07  &  2.85e-04  &  1.38e-03  \\
Sum of Powers 20d &  3.60e-06  & \textbf{ 4.18e-09 } &  1.25e-07  &  4.06e-04  &  2.11e-03  \\
Sum of Powers 50d &  4.24e-06  & \textbf{ 2.92e-09 } &  1.92e-07  &  3.92e-04  &  2.53e-03  \\
Sum of Powers 100d &  4.99e-06  & \textbf{ 9.19e-09 } &  4.53e-04  &  4.20e-04  &  6.37e-03  \\
Sum of Powers 250d &  5.32e-06  & \textbf{ 1.74e-08 } &  1.66e+00  &  3.94e-04  &  3.05e-01  \\
Sum of Powers 500d & \textbf{ 5.63e-06 } &  1.45e-02  &  4.90e+00  &  3.98e-04  &  9.88e-01  \\
Sum of Powers 1000d & \textbf{ 4.38e-06 } &  2.95e-01  &  5.62e+00  &  4.14e-04  &  1.76e+00  \\
\hline
Sum of Squares 2d & \textbf{ 1.76e-12 } &  1.16e-01  &  1.07e-01  &  9.96e-09  &  5.65e-09  \\
Sum of Squares 10d & \textbf{ 2.03e-12 } &  8.17e-11  &  2.58e-10  &  5.69e-07  &  6.02e-07  \\
Sum of Squares 20d & \textbf{ 1.06e-12 } &  4.45e-10  &  1.24e-09  &  2.37e-06  &  2.60e-06  \\
Sum of Squares 50d & \textbf{ 1.83e-12 } &  7.56e-10  &  8.96e-09  &  1.61e-05  &  1.70e-05  \\
Sum of Squares 100d & \textbf{ 1.89e-13 } &  4.80e-09  &  4.70e-08  &  6.49e-05  &  6.73e-05  \\
Sum of Squares 250d &  4.21e-08  & \textbf{ 2.95e-08 } &  3.21e-02  &  4.02e-04  &  4.05e-03  \\
Sum of Squares 500d &  6.24e-05  & \textbf{ 2.96e-07 } &  4.53e-02  &  1.61e-03  &  6.86e-03  \\
Sum of Squares 1000d &  1.25e-01  & \textbf{ 4.90e-06 } &  1.33e+01  &  6.36e-03  &  2.27e+00  \\
\hline
Trid 2d & \textbf{ 5.00e+00 } & \textbf{ 5.00e+00 } & \textbf{ 5.00e+00 } & \textbf{ 5.00e+00 } & \textbf{ 5.00e+00 } \\
Trid 10d & \textbf{ 6.50e+01 } & \textbf{ 6.50e+01 } &  6.51e+01  & \textbf{ 6.50e+01 } & \textbf{ 6.50e+01 } \\
Trid 20d & \textbf{ 2.30e+02 } & \textbf{ 2.30e+02 } &  2.31e+02  &  2.95e+02  & \textbf{ 2.30e+02 } \\
Trid 50d &  1.33e+03  & \textbf{ 1.32e+03 } &  1.35e+03  &  1.47e+04  & \textbf{ 1.32e+03 } \\
Trid 100d &  5.16e+03  & \textbf{ 5.11e+03 } &  1.39e+04  &  1.56e+05  &  1.04e+04  \\
Trid 250d & \textbf{ 5.55e+04 } &  8.92e+05  &  2.18e+06  &  2.61e+06  &  1.57e+06  \\
Trid 500d & \textbf{ 4.67e+06 } &  1.64e+07  &  2.09e+07  &  2.10e+07  &  1.83e+07  \\
Trid 1000d & \textbf{ 9.91e+07 } &  1.57e+08  &  1.68e+08  &  1.67e+08  &  1.62e+08  \\
\hline
Zakharov 2d & \textbf{ 3.61e-12 } &  2.21e-01  &  8.11e-05  &  1.58e-01  &  2.11e-02  \\
Zakharov 10d & \textbf{ 6.40e-12 } &  9.58e-05  &  8.85e-04  &  3.00e+02  &  2.31e+07  \\
Zakharov 20d & \textbf{ 6.28e-12 } &  2.61e-03  &  2.45e+00  &  6.68e+02  &  2.09e+09  \\
Zakharov 50d & \textbf{ 4.78e-12 } &  1.46e+01  &  8.71e+01  &  1.83e+03  &  3.71e+11  \\
Zakharov 100d &  3.20e+06  & \textbf{ 4.54e+01 } &  1.34e+02  &  2.71e+04  &  3.45e+13  \\
Zakharov 250d &  4.94e+11  &  2.12e+02  & \textbf{ 1.89e+02 } &  1.81e+07  &  4.45e+15  \\
Zakharov 500d &  2.82e+14  & \textbf{ 3.46e+02 } &  5.61e+02  &  4.80e+08  &  2.96e+17  \\
Zakharov 1000d &  1.53e+19  &  2.08e+19  &  1.15e+19  & \textbf{ 1.63e+10 } &  2.08e+19 

\end{longtable}

\twocolumn

\clearpage

\end{document}